\documentclass[letterpaper,10pt,journal,twoside]{IEEEtran}
\pdfoutput=1 

\usepackage{amssymb,amsmath}       
\usepackage{array}
\usepackage{bm}                    
\usepackage{epsfig}
\usepackage{float}
\usepackage{graphicx,color,psfrag} 
\usepackage{multirow}              
\usepackage{tabularx}              
\usepackage{wrapfig}

\usepackage[font=footnotesize,skip=5pt]{caption}

\usepackage{tikz}
\usetikzlibrary{automata}
\usetikzlibrary{arrows}
\usetikzlibrary{positioning}
\usetikzlibrary{decorations.pathreplacing}

\usepackage[inline]{enumitem}

\usepackage[ruled,linesnumbered]{algorithm2e}
\let\oldnl\nl
\newcommand{\nonl}{\renewcommand{\nl}{\let\nl\oldnl}}

\usepackage[caption=false,font=footnotesize]{subfig}
\newcommand{\mc}[2][]{{\mathcal{#2}_{\textrm{#1}}}}
\newcommand{\q}[1]{\bm{q}_{\textrm{#1}}}

\newcommand{\T}[1]{\bm{T}_{\textrm{#1}}}
\newcommand{\x}[1]{\bm{x}_{\textrm{#1}}}
\newcommand{\qvect}{\bm{q}}
\newcommand{\Tvect}{\bm{T}}
\newcommand{\GP}{\mathcal{G}\cap\mathcal{P}}
\newcommand{\ie}{{\textit{i.e.}}}
\newcommand{\etal}{\textit{et~al.}}
\newcommand{\tr}[1]{\textrm{#1}}

\newcommand{\start}{\textrm{start}}
\newcommand{\goal}{\textrm{goal}}
\newcommand{\func}[1]{{\tt{#1}}}
\newcommand{\kw}[1]{{\tt{#1}}}

\newtheorem{remark}{Remark}

\newcommand{\comment}[1]{}

\renewcommand{\tt}{\fontfamily{cmtt}\selectfont}

\graphicspath{{figures/}}

\IEEEoverridecommandlockouts

\title{A Single-Query Manipulation Planner}
\author{Puttichai Lertkultanon and Quang-Cuong Pham%
  \thanks{This work was partially supported by Tier $1$ grant
    RG$109/14$ awarded by Singapore MOE to the second author.}
  \thanks{Puttichai Lertkultanon and Quang-Cuong Pham are with the
    School of Mechanical and Aerospace Engineering, Nanyang
    Technological University, $50$ Nanyang Avenue, Singapore $639798$
    {\tt\small L.Puttichai@gmail.com}}%
}

\date{}

\begin{document}
\maketitle

\setlength{\algomargin}{2em}
\SetKwIF{If}{ElseIf}{Else}{if}{ then}{elif}{else}{}%
\SetKwFor{For}{for}{ do}{}%
\SetKwFor{ForEach}{foreach}{ do}{}%
\SetKwFor{While}{while}{ do}%
\AlgoDontDisplayBlockMarkers\SetAlgoNoEnd\SetAlgoNoLine%
\DontPrintSemicolon

\begin{abstract}
  In manipulation tasks, a robot interacts with movable object(s). The
  configuration space in manipulation planning is thus the Cartesian
  product of the configuration space of the robot with those of the
  movable objects. It is the complex structure of such a ``composite
  configuration space'' that makes manipulation planning particularly
  challenging. Previous works approximate the connectivity of the
  composite configuration space by means of discretization or by
  creating random roadmaps. Such approaches involve an extensive
  pre-processing phase, which furthermore has to be re-done each time
  the environment changes. In this paper, we propose a high-level
  Grasp-Placement Table similar to that proposed by Tournassoud et al.
  (1987), but which does not require any discretization or heavy
  pre-processing. The table captures the potential connectivity of the
  composite configuration space while being specific only to the
  movable objects: in particular, it does not require to be
  re-computed when the environment changes. During the query phase,
  the table is used to guide a tree-based planner that explores the
  space systematically. Our simulations and experiments show that the
  proposed method enables improvements in both running time and
  trajectory quality as compared to existing approaches.
\end{abstract}


\section{Introduction}
\label{section:introduction}

Automated robotic assembly requires the ability to plan
\emph{pick-and-place} motions\,: pick an object from a given
configuration (position and orientation) and place it in a desired
configuration, possibly with a desired grasp. A grasp refers to a
relative transformation between the gripper and the object. When none
of the possible grasps for picking up the object at its initial
configuration is compatible with subsequent operations, the robot
needs to change its grasp, possibly several times, until the desired
grasp can be attained. Unlike multi-fingered hands, parallel grippers
(which are the most common and robust grippers in the industry) cannot
realize in-hand manipulations. To change a grasp with a parallel
gripper, the robot must \emph{transfer} the object to some
intermediate stable placement, then \emph{transit} to a new
grasp. Those operations are collectively termed
``regrasping''~\cite{TP87icra}.

Since a manipulation planning problem involves both the robot and the
object configurations, searching for a sequence of transit and
transfer paths, so-called a \emph{manipulation path}, has to be done
in the \emph{composite configuration space} $\mc{C}$~\cite{SimX04ijrr},
which is the Cartesian product of the robot configuration space
$\mc[robot]{C}$ and the object configuration space $\mc[object]{C}$. A
transit path, along which the robot moves alone while the object
remains at a stable placement, lies in $\mc{P}$, the subset of
$\mc{C}$ corresponding to stable placements of the object. Similarly, a
transfer path, along which the robot moves while grasping the object
with a valid grasp, lies in $\mc{G}$, the subset of $\mc{C}$
corresponding to valid grasps. The subset $\GP$ is central in solving
manipulation planning problems since it is the place where transitions
between transit and transfer paths may
occur~\cite{ALS94wafr},~\cite{SimX04ijrr}. Manipulation planners
differ mainly in the way they explore and capture the connectivity of
$\GP$.

For a parallel gripper grasping a polyhedral object, all the valid
grasps and stable placements can be categorized into a finite number
of classes. A grasp class corresponds to e.g. a pair of object
surfaces the fingers are touching. A placement class corresponds to
e.g. a surface (of the object or of its convex hull) in contact with
the table. Each grasp and placement class may be further parameterized
by a set of continuously varying parameters.

In a pioneering work, Tournassoud \etal~\cite{TP87icra} started by
discretizing $\GP$\footnote{The notion of composite configuration
  space was not introduced at the time. However, their idea of grasp
  and placement spaces bears a close resemblance to the formulation
  presented in~\cite{SimX04ijrr}.}  and then searched for a feasible
regrasping sequence by backward chaining from the goal grasp. However,
because of the high dimensionality of $\mc{C}$ and the complexity
arising from the discretization, the authors had to constrain each
grasp and placement class to have only single varying parameter. Their
method is therefore limited and much dependent on the particular
choice of the parameters. Nevertheless, they did introduce the
important notion of \emph{Grasp-Placement Table}, which captures part
of the connectivity of $\GP$. This Grasp-Placement Table can be seen
as an instance of a Manipulation Graph~\cite{ALS94wafr} that is
particularly adapted for polyhedral objects. The Grasp-Placement Table
is a grid where each vertical line represents a placement class while
each horizontal line represents a grasp class. Intersections of
vertical lines with horizontal lines then correspond to subsets of
$\GP$. A transfer path appears as a connection between intersections
on the same horizontal line, whereas a transit path appears as a
connection between intersections on the same vertical line.

Here we propose a method to construct a \emph{high-level}
Grasp-Placement Table (or graph). In contrast with~\cite{TP87icra},
our graph does not require any discretization or heavy
pre-processing. Moreover, it is specific only to the movable object,
and not to the environment or to the robot\,: it does not therefore
require to be re-computed when the environment changes. At the query
phase, the graph is used as a high-level task planner to guide the
manipulation planner in exploring $\mc{C}$.

Specifically, the edges of the graph are generated by ignoring all the
robot kinematics. Verification of kinematic feasibility along each
edge of the graph is postponed to the planning phase. In doing so, our
method enables handling the full parameterization of $\GP$. Although
the idea of delaying IK computation has been independently explored in
a recent work~\cite{WanX15icra} to construct a manipulation graph, the
authors' graph still requires discretizing grasp and placement
classes, entailing the same problem of heavy pre-processing.

By constructing and using a high-level Grasp-Placement Table, we
decouple a pick-and-place manipulation planning problem into two
layers of planning. The \emph{high-level} task planning layer consists
in finding a sequence of $\GP$ configurations that answers the query,
while the \emph{low-level} motion planning layer consists in finding
\emph{actual} motions between the $\GP$ configurations. Since the
emphasis of this paper is on the high-level planning, \ie, how we
construct and use a high-level Grasp-Placement Table to help solve a
manipulation query, addressing uncertainty, which takes place at the
second layer, is beyond the scope of this paper.

Note also that we focus here on industrial assembly, where
parallel-jaw grippers are pervasive owing to their cost-effectiveness
and ease of integration. We first specifically consider the case when
movable objects are either boxes or composed of boxes. These
properties enables an efficient parameterizations of grasps and
placements. We discuss extension to broader types of objects in
Section~\ref{subsection:extension}.

The rest of this paper is organized as follows. In
Section~\ref{section:related_lit}, we briefly review related
literature. In Section~\ref{section:notation}, we present definitions
and conventions that will be used in the
sequel. Section~\ref{section:graph} introduces the high-level
Grasp-Placement Table, its construction, and its use in
planning. Comparisons between the proposed planner and other
manipulation planners are presented in
Section~\ref{section:results}. Finally,
Section~\ref{section:discussion} offers a brief discussion and
sketches future research directions.

\section{Related Literature}
\label{section:related_lit}

\subsection{Manipulation Planning}
Manipulation planners can be seen as a generalization of motion
planners where the robot is allowed to displace specific objects,
called movable objects, in its environment via specific interactions,
e.g., pushing or grasping. Early work considered pick-and-place
manipulation planning as a \emph{regrasping
  problem}~\cite{TP87icra},~\cite{RW97icra},~\cite{StoX99icra}. The
authors discretized $\GP$ and checked at each discretized point
whether the placement was stable and the grasp was feasible. They then
executed a deterministic search of a regrasping sequence on the set of
feasible placements and grasps. Evaluation of object placements based
on minimum tipping energy was proposed in~\cite{RW97icra}.

A recent work on regrasping algorithms~\cite{WanX15icra} also utilized
a discretization of $\GP$ to construct a regrasp graph, which
represents connectivity of $\GP$ connected components, to analyze
workcell utility. The authors improved efficiency of their algorithm
by delaying IK computation in the graph construction until it is
necessary. However, the discretization cost still remains considerable
compared to our method of construction.

Most later work is based on the notion of Manipulation
Graph~\cite{ALS94wafr} and composite configuration
space~\cite{SimX04ijrr}. Initially the planners usually assumed
discrete sets of grasps and
placements~\cite{ALS94wafr},~\cite{KL94icra},~\cite{NK00iros}. Later
work extended the approach to handle continuous
representation~\cite{SimX04ijrr}. The approach has also been extended
to facilitate bimanual manipulation planning~\cite{HarX14icra}. Apart
from permitting only two \emph{modes} of motions, transit and
transfer, there is also work on generalization that permitting finding
solution paths with multiple modes of
motions~\cite{HN11ijrr},~\cite{BarX13er}.

Although a majority of manipulation planners are sampling-based, there
also exist deterministic manipulation
planners~\cite{CCL10icra},~\cite{CCL13ijrr}. These planners construct
a lattice graph representing a discretized configuration space and
then use heuristic searches to find solution paths. However, these
planners are currently capable of planning only single-mode motions
such as reach-to-grasp motions or motions once the robot has already
grasped an object.

\subsection{Object Rearrangement}
Another problem related to manipulation planning is object
rearrangement problem~\cite{StiX07icra},~\cite{KB15rss}. The problem
consists in finding a rearrangement path such that all movable objects
are moved by the robot to their goal poses. Generally the problem can
be decomposed into two subproblems. The first problem is finding a
sequence of object rearrangements and the second one is finding
manipulation paths connecting adjacent rearrangements. Here
pick-and-place planners can serve as local planners to connect nodes
of object rearrangements.

\subsection{Task and Motion Planning}
Planning pick-and-place motions can also be considered in a framework
of task and motion planning~\cite{CAG09ijrr}, \cite{PK14iros},
\cite{SriX14icra}, \cite{GPK15afr}. A planning problem is solved
through two layers of planning\,: high-level task planning and
low-level motion planning. A task planner executes symbolic searches
for high-level actions such as \emph{pick}, \emph{move}, and
\emph{place}, not considering geometries nor kinematics. A motion
planner then computes actual commands to follow the strategy---a
sequence of actions---given by the task planner.

From the perspective of task and motion planning, instead of planning
tasks by a symbolic task planner, we generate task plans by using
information provided by our high-level Grasp-Placement Table. A grasp
search algorithm is used to extract task plans, which will then inform
the manipulation planner about how to make transitions between $\GP$
connected components. This makes the overall planner less likely to
generate solutions with redundant grasp and ungrasp operations.

\section{Definitions and Convention}
\label{section:notation}

In this section, we present the definitions and conventions involved
in manipulation planning. The terminology introduced here will
facilitate the discussions in the sequel.

\subsection{Mathematical Definitions}
\label{subsection:definitions}
We represent a composite configuration with a couple $(\qvect,
\Tvect)$, where $\qvect \in \mc[robot]{C} \subseteq \mathbf{R}^{n}$ is
a robot configuration, $n$ is the degrees-of-freedom (DOF) of the
robot, and $\Tvect \in SE(3)$\footnote{$SE(3)$ is the special
  Euclidean group of rigid body transformations.} is a homogeneous
transformation matrix of the object. For a composite configuration in
$\mc{G}$ we define a grasp as a vector of parameters describing the
relative transformation of the gripper and the object. Therefore, a
grasp can generally be described by a vector of $7$ parameters of
quaternion and translation, which can be uniquely identified from
$\qvect$ and $\Tvect$. The number of parameters, however, can be
reduced via grasp parameterization. An example of grasp
parameterization can be found in~\cite{YN10robio}.

A \textbf{single-mode path} is defined as a continuous function $P$
from the unit interval $[0, 1]$ to a level set of $\mc{G}$ or
$\mc{P}$. There are two types of single-mode paths\,: transit and
transfer. A \textbf{transit path} maps the unit interval into a level
set of $\mc{P}$, \ie, for any two configurations $(\qvect_{1},
\Tvect_{1})$ and $(\qvect_{2}, \Tvect_{2})$ on the path, $\Tvect_{1} =
\Tvect_{2}$. A \textbf{transfer path} maps the unit interval into a
level set of $\mc{G}$, \ie, a grasp remains constant along the path.

Next, we define a binary operation called \textbf{composition}
operation. First of all, the composition of two single-mode paths
$P_{1}$ and $P_{2}$ is defined as
\begin{equation*}
  (P_{1} \ast P_{2})(s) = \left\{  
    \begin{matrix}
      P_{1}(2s/t) & 0 \leq s \leq t/2,\\
      P_{2}(2s/t - 1) & t/2 \leq s \leq t,
    \end{matrix}
  \right.
\end{equation*}
where $t = 1$ if both paths are of the same type and $t = 2$
otherwise. Note that the composition operation is only defined when
$P_{1}(1) = P_{2}(0)$. From the above definition, when $P_{1}$ and
$P_{2}$ have the same type, $M = P_{1} \ast P_{2}$ is also a
single-mode path. We use $|M|$ to denote the \textbf{parameterization
  domain length} or \textbf{domain length} of $M$. Note that all
single-mode paths have unit domain length.

Let $\prod_{i = a}^{b}P_{i} = P_{a} \ast P_{a + 1} \ast \ldots \ast
P_{b}$, where $b \geq a$. We can define the composition of $k$
single-mode paths as
\begin{equation*}
  \left( \prod_{i = 1}^{k} P_{i} \right)(s) 
  = \left\{
    \begin{matrix}
      (P_{1} \ast P_{2})(s) & 0 \leq s \leq t_{2}\\
      M_{2}(s - t_{2}) & t_{2} \leq s \leq |M_{2}| + t_{2}
    \end{matrix}
  \right.
\end{equation*}
\begin{equation*}
  = \left\{
    \begin{matrix}
      M_{1}(s) & 0 \leq s \leq |M_{1}|\\
      (P_{k - 1} \ast P_{k})(s - |M_{1}|) & |M_{1}| \leq s \leq
      |M_{1}| + t_{1},
    \end{matrix}
  \right.
\end{equation*}
where $M_{1} = \prod_{i = 1}^{k - 2}P_{i}$, $M_{2} = \prod_{i =
  3}^{k}P_{i}$, and $t_{1}$ and $t_{2}$ are one if both single-mode
paths are of the same type and two otherwise. Notice that the
composition operation is associative but not commutative.

According to the above definition of the composition operation, we can
now define a \textbf{manipulation path} as a composition of
single-mode paths. A manipulation path is \textbf{irreducible} if no
two consecutive single-mode paths are of the same type, \ie, it is an
alternating composition of transit and transfer paths. Any composition
$\prod_{i = 1}^{k}P_{i}$ can always be written it as an irreducible
manipulation path $\prod_{i = 1}^{l}P_{i}'$ with $l \leq k$. For an
irreducible manipulation path $M$, we have that the number of
\textbf{transitions} (between transit and transfer paths or vice
versa) is $|M| - 1$.

\begin{figure}
  \centering
  \begin{tikzpicture}
    \node at (0, 0) {\includegraphics[width = 0.1\textwidth] {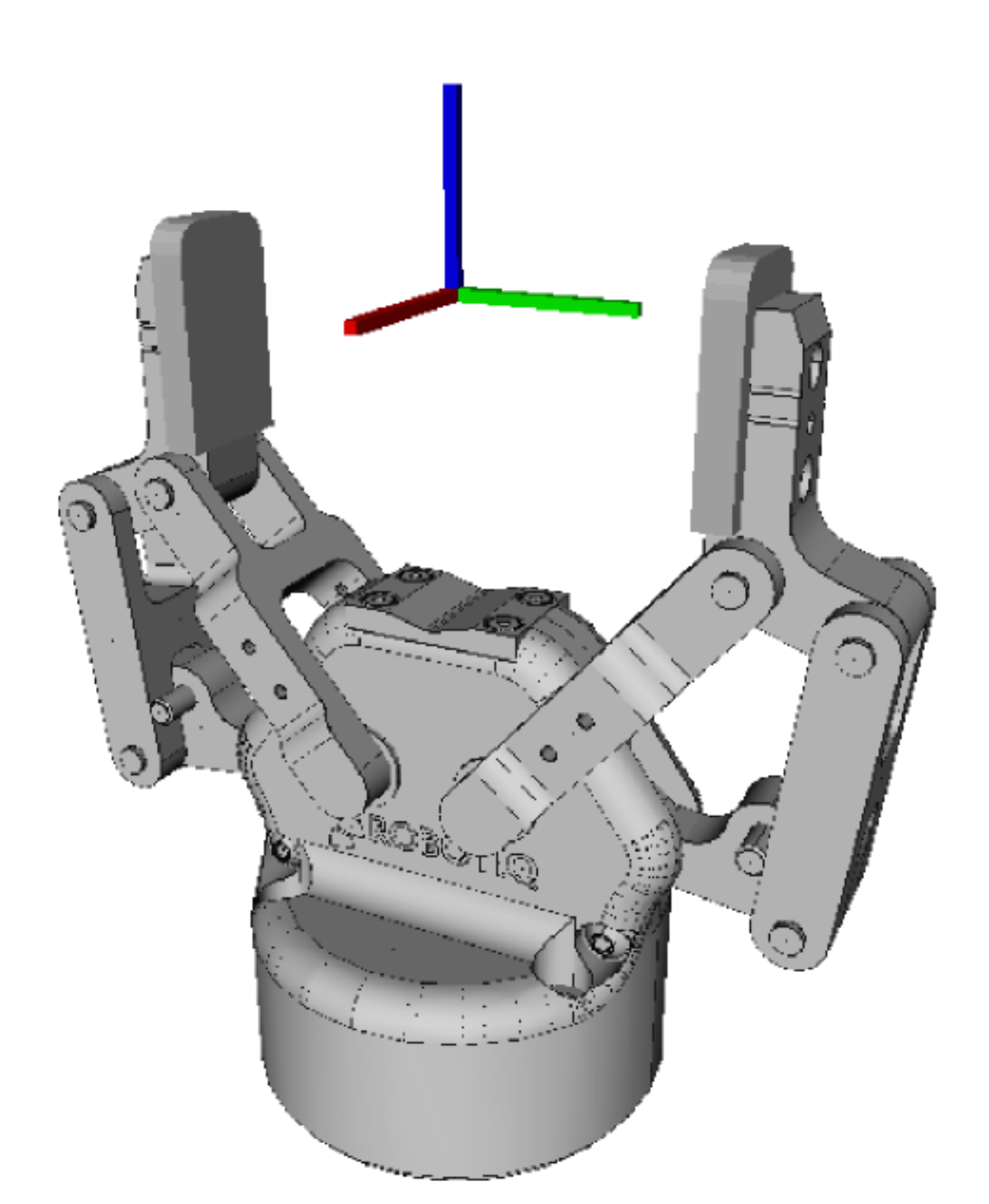}};
    \node [scale=0.8] at ( 1.4, 0.4) {lateral};
    \node [scale=0.8] at (-1.4, 0.4) {sliding};
    \node [scale=0.8] at ( 0, 1.2) {approaching};
  \end{tikzpicture}
  \caption{The parallel gripper used in this paper with its local
    frame. The lateral direction is orthogonal to both finger
    surfaces. The sliding direction is parallel to both finger
    surfaces and is defined such that the approaching direction is
    pointing out of the gripper.}
  \label{figure:gripper}
\end{figure}
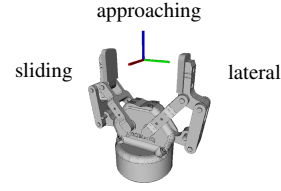

\subsection{Placement Classes and Grasp Classes}
The sets $\mc{P}$ and $\mc{G}$ can be partitioned into finite disjoint
classes called placement classes and grasp classes, respectively. A
class groups together composite configurations with similar
properties. Each placement class indicates how an object is placed on
a table, and thus normally refers to a surface of the object's convex
hull being in contact with the table. Similarly, each grasp class
indicates how the object is grasped by the gripper. In our case, with
an assignment of the gripper's local frame as shown in
Fig.~\ref{figure:gripper}, each grasp class refers to the relative
direction of the gripper's approaching direction with respect to the
object.

Consider a gripper grasping a box. We say that the approaching
direction is $+x$ if the positive direction of the gripper's
approaching axis is aligned with $+x$-direction of the box's local
frame. Therefore, there are $6$ possible directions along which the
gripper can approach the box\footnote{This means the approaching axis
  of the gripper must be perpendicular to a surface of the box. A
  configuration that the approaching direction is not perpendicular to
  any surface can then be achieve by rotating the gripper about the
  sliding axis.}. For convenience, we will use integers $1$ to $6$ to
denote approaching directions $+x, +y, +z, -x, -y, -z$,
respectively. Now consider the case when the object is composed of $m$
boxes. Suppose we index those boxes with integers from $1$ to
$m$. There will be in total $6m$ possible grasp classes. When the
gripper is approaching the object in the direction $i$ of the box $j$,
the grasp class index is $i + 6(j - 1)$. For example, in the grasp
class $7$, the gripper is grasping box $2$ and the approaching axis is
aligned with $+x$-axis of the box.

\section{Manipulation Planning Using High-Level Grasp-Placement Table}
\label{section:graph}

\subsection{High-Level Grasp-Placement Table}

\subsubsection{Overview}
\label{subsubsection:graph_overview}
A high-level Grasp-Placement Table (or graph) is an undirected,
unweighted graph whose nodes represent different subsets of
$\GP$. According to the partitioning of $\mc{G}$ and $\mc{P}$, the set
$\GP$ is then partitioned into finite disjoint subsets. Each subset of
$\GP$ is the intersection between a particular pair of placement and
grasp classes. Therefore, a node in the graph can be represented by a
pair of integers, a placement class index and a grasp class index. We
visualize a high-level Grasp-Placement Table in the similar way as the
authors of~\cite{TP87icra} did for their Grasp-Placement Table. Our
Table is plotted on a two-dimensional grid. Each vertical line
corresponds to a placement class whereas each horizontal line
corresponds to a grasp class. Intersections of vertical with
horizontal lines then represent subsets of $\GP$.

If there were no collision avoidance constraints and kinematic
reachability constraints, any combination of an object contact surface
and a gripper's approaching direction would have been possible. Any
pair of nodes on the same vertical or horizontal line will be
connected. Imposing those constraints makes some nodes and edges
infeasible. However, verifying all constraints at the graph
construction phase entails costly computations. Also, by checking
kinematic reachability, the graph will become robot- and
environment-dependent\,: it will need to be re-computed when the
environment changes. Therefore, we propose to construct the graph by
verifying only collision avoidance constraints (between the gripper
and the table). Robot kinematic reachability constraint verification
(IK computation) is postponed until the planning phase.

\subsubsection{Graph Construction}
\label{subsubsection:graph_construction}
In the first step, we check for feasibility of graph nodes. For each
placement class, we place the object at a nominal location on the
table. Then for each box composing the object, we check whether the
gripper can approach the box from any of the $6$ directions, \ie,
$+x$-, $+y$-, $+z$-, $-x$-, $-y$-, and $-z$-directions, of the box's
local frame, without colliding with the table. If there is no
collision, we add a node that represents the respective placement and
grasp classes into the graph. We continue this procedure for all
placement classes. Since we do not consider the robot kinematics here,
the actual location of the object on the table, \ie, how far the
object is from the robot, does not affect the resulting graph.

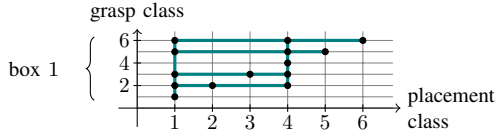
\begin{figure}
  \centering
  \begin{tikzpicture}[scale=0.5]
    \def\np{6}
    \def\ng{6}
    \def\ystep{0.3}
    \draw[ystep = \ystep cm, gray, ultra thin](0, 0) grid 
    (\np + 0.8, \ng*\ystep + 0.8*\ystep);
    \draw[->] (-0.5, 0) -- (\np + 1, 0);
    \draw[->] (0, -0.5) -- (0, \ng*\ystep + \ystep);
    \node [align=left, right, scale=0.8] at (\np + 1, 0) {placement\\
      class};
    \foreach \x in {1, ..., \np}{
      \draw (\x, -0.1) -- (\x, 0.1);
      \node [below, scale=0.8] at (\x, 0) {$\x$};
    }
    \node [align=left, above, scale=0.8] at (0, \ng*\ystep + \ystep)
    {grasp class};
    \foreach \y in {2, 4, ..., \ng}{
      \draw (-0.1, \y*\ystep) -- (0.1, \y*\ystep);
      \node [left, scale=0.8] at (0, \y*\ystep) {$\y$};
    }

    \draw[very thick, teal] (1, 1*\ystep) -- (1, 6*\ystep);
    \draw[very thick, teal] (4, 2*\ystep) -- (4, 6*\ystep);
    \draw[very thick, teal] (1, 2*\ystep) -- (4, 2*\ystep);
    \draw[very thick, teal] (1, 3*\ystep) -- (4, 3*\ystep);
    \draw[very thick, teal] (1, 5*\ystep) -- (5, 5*\ystep);
    \draw[very thick, teal] (1, 6*\ystep) -- (6, 6*\ystep);
    \foreach \x/\y in 
    { 1/1, 1/2, 1/3, 1/5, 1/6,
      2/2,
      3/3, 
      4/2, 4/3, 4/4, 4/5, 4/6,
      5/5,
      6/6 }{
      \filldraw (\x, \y*\ystep) circle [radius=0.08cm, fill=black];
    }
    \draw [decorate, decoration={brace, amplitude=3pt}, xshift=-4pt,
    yshift=0pt] (-1, 0.7*\ystep + 6*\ystep - 6*\ystep) -- 
      (-1, 0.3*\ystep + 6*\ystep)
    node [black, midway, xshift=-0.8cm] {\footnotesize box $1$};
    
  \end{tikzpicture}
  \caption{An example of a high-level Grasp-Placement Table for a box
    of dimension $28.0$ cm. $\times$ $4.9$ cm. $\times$ $2.5$ cm. For
    clarity, self-loops are not depicted. Note that every node on the
    same line is connected to all other nodes although we do not draw
    separate lines for them. For example, node $(1, 6)$ is
    reachable from $(6, 1)$ in one step.}
  \label{figure:graph_box}
\end{figure}

After we obtain all feasible nodes in the first step, we display them
on a grid. Then we connect every pair of nodes on the same vertical or
horizontal line. These edges represent \emph{potential} connections
between nodes. An edge connecting a pair of nodes on the same vertical
line represents \emph{transit} paths while an edge connecting a pair
of nodes on the same horizontal line represents \emph{transfer}
paths. Every node also has two self-loops. One loop corresponds to a
transit path; the other loop corresponds to a transfer
path. Fig.~\ref{figure:graph_box} shows an example of a high-level
Grasp-Placement Table for a box of dimension $28.0$ cm. $\times$ $4.9$
cm. $\times$ $2.5$ cm. For simplicity, we do not show self-loops in
the plot.

The construction can be generalized to handle start and goal
configurations in $\mc{G}$ and/or $\mc{P}$ by adding special nodes
into the graph. To handle configurations in $\mc{P}$, we can add a
grasp class index $0$ to the graph. For example, in the above example
of the box, we will obtain $6$ new nodes, $(1, 0), (2, 0), \ldots, (6,
0)$. These nodes are connected to all other nodes with the same
placement class. Note, however, that there are no horizontal edges
between them since it is not possible to travel directly from one
placement class to another in $\mc{P}$. To handle configurations in
$\mc{G}$, we can add a placement class index $0$, similarly to the
previous case.

\subsubsection{Guiding a Manipulation Planner via Task Plans}
\label{subsubsection:extractingplans}
One of quality measures of manipulation paths is the number of
transitions (as defined in
Section~\ref{subsection:definitions}). Fewer transitions means fewer
grasp/ungrasp operations and fewer single-mode paths. This may
therefore lead to a shorter overall execution time.

From the constructed graph we can extract \emph{task plans} of a
specific length\footnote{A task plan of length $k$ has $k - 1$
  transitions} which serve as a guide for the planner about how to
travel between different subsets of $\GP$. The planner, instead of
exploring randomly how to go from one subset of $\GP$ to another, will
follow those plans to search for a manipulation path of a specific
number of transitions. Here a \textbf{task plan} is a sequence of
graph nodes connected by graph edges. For example, from
Fig.~\ref{figure:graph_box} a task plan of length $3$ to travel from
$(6, 6)$ to $(2, 2)$ is given by $(6, 6) \rightarrow (4, 6)
\rightarrow (4, 2) \rightarrow (2, 2)$.

The task plans extracted from the graph only provide high-level
information on how to pick and place the object. Exact parameter
values for each grasp and placement classes will be assigned by the
planner, \ie, via random sampling, in the planning phase (see
Section~\ref{subsection:algorithm}).

\subsection{Manipulation Planning Algorithm}

\subsubsection{Algorithm Details}
\label{subsection:algorithm}
Our pick-and-place manipulation planner proceeds in two phases\,:
pre-processing phase and planning phase. In the pre-processing phase,
it constructs a high-level Grasp-Placement Table based on the models
of the object and of the gripper (the latter is needed for collision
checking). Note that neither the kinematic model of the robot nor the
environment is needed at this phase. Then this high-level
Grasp-Placement Table is used in the planning phase to guide a
bidirectional tree-based planner, similar to~\cite{LK00citeseer}, to
search for a manipulation path. The main algorithm is listed in
Algorithm~\ref{algo:main}.

\begin{algorithm}[h]
\caption{Main algorithm}
\label{algo:main}
\Indm
{\nonl{\func{Main}($\mc{R}, \mc{M}, \q{start}, \T{start}, \q{goal}, 
    \T{goal}, t_{\textrm{max}}$):}}\;
\Indp
$G \leftarrow$ \func{Preprocess}($\mc{R}, \mc{M}$)\;
$v_{\start} \leftarrow$ \func{Vertex}($\q{start}, \T{start}$)\;
$v_{\goal} \leftarrow$ \func{Vertex}($\q{goal}, \T{goal}$)\;
$l \leftarrow$
\func{FindShortestPathLength}($G, v_{\start}, v_{\goal}$)\;
\label{line:findshortestlength}
$t \leftarrow 0, k \leftarrow l$, 
found $\leftarrow$ \kw{False}\;
$\Delta \leftarrow$ \func{ChoosePathLengthIncrement}$(v_{\start}, v_{\goal})$\;
\While{\upshape ($t < t_{\textrm{max}}$) \kw{and} (\kw{not} found)}{
  $t_{\textrm{s}} \leftarrow$ \func{GetTime}()\; 
  $\Pi \leftarrow$ \func{FindPlansOfGivenLength}($k, v_{\start},v_{\goal}$)\;
  \label{line:findplansofgivenlength}
  $Q \leftarrow$ \func{CreateDirectedGraph}($\Pi$)\;
  \{found, $\pi$\} $\leftarrow$ \func{PlanPath}($Q, \mc{R}, \mc{M}, v_{\start},
  v_{\goal}, t_{\textrm{max}} - t$)\;
  $t_{\textrm{e}} \leftarrow$ \func{GetTime}()\;
  $t \leftarrow t + (t_{\textrm{e}} - t_{\textrm{s}})$\;
  \If{\upshape found}{
    \KwRet{$\pi$}
  }
  $k \leftarrow k + \Delta$\;
}
\KwRet{\upshape \kw{None}}\;
\end{algorithm}

The main algorithm takes as its input a robot model $\mc{R}$; an
object model $\mc{M}$; start and goal robot configurations $\q{start}$
and $\q{goal}$; start and goal object transformations $\T{start}$ and
$\T{goal}$; and a maximum running time for the planner
$t_{\textrm{max}}$. Details of functions in Algorithm~\ref{algo:main}
are listed below\,:

\begin{itemize}[leftmargin=*]
\item \func{Preprocess} gathers geometric information of the gripper
  and the object to construct a high-level Grasp-Placement Table $G$,
  as described in Section~\ref{subsubsection:graph_construction}.
  
\item \func{Vertex} initializes a new tree vertex to store information
  of a composite configuration. Note that this function also examines
  $G$ to find the graph node to which the input composite
  configuration belongs.

\item \func{FindShortestPathLength} finds the length $l$ of the
  shortest task plan(s) to go from the graph node $c_{\start}$ to the
  graph node $c_{\goal}$. Here $l - 1$ serves as the lower bound of
  the number of transitions of manipulation paths connecting
  $(\q{start}, \T{start})$ and $(\q{goal}, \T{goal})$.

\item \func{ChoosePathLengthIncrement} chooses the suitable value of
  $\Delta$ for the query. If either $(\q{start}, \T{start})$ or
  $(\q{goal}, \T{goal})$ is in $\GP$, $\Delta$ can be $1$. Otherwise,
  $\Delta$ has to be $2$ in order to keep the resulting manipulation
  path irreducible.

\item \func{CreateDirectedGraph} creates a directed graph $Q$ from
  task plans in $\Pi$. A node of $Q$ is encoded as a couple $(d_{i},
  c_{i})$, where $c_{i}$ is the corresponding node of $G$ and $d_{i}$
  indicates the number of steps $c_{i}$ is away from
  $c_{\start}$\footnote{We need to encode the level $d_{i}$ of $c_{i}$
    into each node of $Q$ since the node $c_{i}$ may appear at
    different steps in different task plans. This helps the planner
    not to get lost when many task plans contain the same node
    $c_{i}$.}.

\item \func{PlanPath}, which acts as a motion planner, searches for a
  manipulation path according to information provided by
  $Q$. \func{PlanPath} is listed in Algorithm~\ref{algo:planpath}.
\end{itemize}

\vspace{-5pt}
\begin{algorithm}[h]
\caption{Planning phase}
\label{algo:planpath}
\Indm
{\nonl{\func{PlanPath}($Q, \mc{R}, \mc{M}, v_{\start}, v_{\goal}, 
    t_{\textrm{max}}$):}}\;
\Indp
$\mc[FW]{T} \leftarrow$ \func{Tree}($v_{\start}$),
$\mc[BW]{T} \leftarrow$ \func{Tree}($v_{\goal}$)\;
$\mc[a]{T} \leftarrow \mc[FW]{T}$,
$\mc[b]{T} \leftarrow \mc[BW]{T}$\; $t \leftarrow 0$\;
\While{\upshape $t < t_{\textrm{max}}$ \kw{or} $Q$.haspath}{
  $t_{\textrm{s}} \leftarrow$ \func{GetTime}()\;
  $v \leftarrow$ \func{SampleTree}($\mc[a]{T}$)\;
  $Q \leftarrow$ \func{RemoveInfeasibleEdges}($Q$)\;
  result $\leftarrow$ \func{ExtendFrom}($v, Q$)\;
  \If{\upshape result == \kw{REACHED}}{
    $\pi \leftarrow$ \func{ExtractPath}($\mc[FW]{T}, \mc[BW]{T}$)\;
    \KwRet{\upshape\{\kw{True}, $\pi$\}}
  }
  $t_{\textrm{e}} \leftarrow$ \func{GetTime}()\;
  $t \leftarrow t + (t_{\textrm{e}} - t_{\textrm{s}})$\;
  \func{Swap}($\mc[a]{T}, \mc[b]{T}$)\;
}
\KwRet{\upshape\{\kw{False}, \kw{None}\}}
\end{algorithm}
\vspace{-5pt}

\begin{itemize}[leftmargin=*]
\item \func{ExtendFrom} first randomly samples a new composite
  configuration in the grasp and placement classes suggested by
  $Q$. Then it will attempt to connect the composite configuration
  contained in $v$ with the newly sampled one. The function returns
  \kw{True} if the attempt is successful.

\item \func{RemoveInfeasibleEdges} removes edges in $Q$ which lead to
  more than $N$ failed attempts by \func{ExtendFrom}, where $N$ is a
  threshold value set by the user.
\end{itemize}

\subsubsection{Example}
\label{subsubsection:example}
Consider the task of moving the box, whose high-level Grasp-Placement
Table is shown in Fig.~\ref{figure:graph_box}, from the placement and
grasp $c_{\start} = (6, 6)$ to $c_{\goal} = (2, 2)$. In the first
planning loop in Algorithm~\ref{algo:main}, the algorithm will try to
find a manipulation path of length $k = 3$. The set $\Pi$ will store
two task plans\,: $(6, 6) \rightarrow (4, 6) \rightarrow (4, 2)
\rightarrow (2, 2)$ and $(6, 6) \rightarrow (1, 6) \rightarrow (1, 2)
\rightarrow (2, 2)$. The directed graph $Q$, constructed from $\Pi$,
will serve as a guidance for the planner as follows.

In each while loop in Algorithm~\ref{algo:planpath}, the planner will
randomly pick an existing vertex $v_{\tr{sample}}$ on a tree. Suppose
$v_{\tr{sample}}$ contains a composite configuration in placement and
grasp classes $(4, 6)$. From $Q$, the planner then knows that the next
extension should be attempted towards a composite configuration in the
placement and grasp classes $(4, 2)$. \func{ExtendFrom} will then
sample a composite configuration in that subset of $\GP$ and call a
local planner, e.g., a bidirectional RRT planner~\cite{LK00citeseer},
to attempt the connection.

\subsection{Remarks}
\begin{remark}
  The high-level Grasp-Placement Table $G$, and hence $Q$, is
  constructed without considering kinematic constraints. Failed
  attempts by \func{ExtendFrom} are likely due to kinematic
  infeasibility. Therefore, we need to set a threshold $N$ to prevent
  the planner from repetitively attempting infeasible connections.

  This threshold also affects the running time of our planner. If we
  set $N$ too low, feasible edges of $Q$ can also be removed due to
  false negative reports of kinematic infeasibility. This will lead to
  obtaining solutions with redundant transitions. On the other hand,
  if the threshold is set too high, it will take longer time for the
  planner to declare an infeasible edge and hence longer overall
  running time.
\end{remark}

\begin{remark}
  We can tailor the sampling of vertices from a tree, done in
  \func{SampleTree}, by putting different sampling weight on different
  vertices. For example, we can use weights proportional to the level
  of $c_{\tr{sample}}$ in $Q$, \ie, $d_{\tr{sample}}$. In this way,
  vertices farther away from the root will be more likely to be
  selected.
\end{remark}

\begin{remark}
  Our planner tends to work better in a less constrained
  environment. In a very constrained environment\footnote{This may be
    seen as a narrow passage in $\mc{C}$. An example of such cases is
    the problem considered in~\cite{SimX04ijrr} where the robot needs
    to pull a bar out of a tight cage mounted on the floor.} in which
  the robot needs to grasp and ungrasp the object a number of times
  before it can move the object to the final transformation, our
  planner will spend a considerable amount of time verifying
  infeasible edges. However, in actual industrial settings, although
  the environment might be cluttered, it is usually not tightly
  constrained.
\end{remark}

\section{Results and Comparisons with Other Manipulation Planners}
\label{section:results}

In this section, we present details of comparisons between our planner
and two other planners\,: Primitive Manipulation Planner
(Section~\ref{subsubsection:pmp}) and Discretization-Based
Manipulation Planner (Section~\ref{subsubsection:dbmp}). We
implemented all planners in Python and used OpenRAVE~\cite{Dia10these}
as a simulation environment. The local planner employed in all
manipulation planners was the OpenRAVE built-in bidirectional RRT. The
robot was a $6$-DOF industrial manipulator Denso VS-$060$ equipped
with a $2$-finger Robotiq gripper $85$. All simulations were run on a
$3.2$ GHz Intel\textsuperscript{\circledR} Core\texttrademark desktop
with $3.8$ GB RAM.

\subsection{Task Details}
\begin{figure}
  \centering
  \subfloat[]{\includegraphics[width=0.13\textwidth]{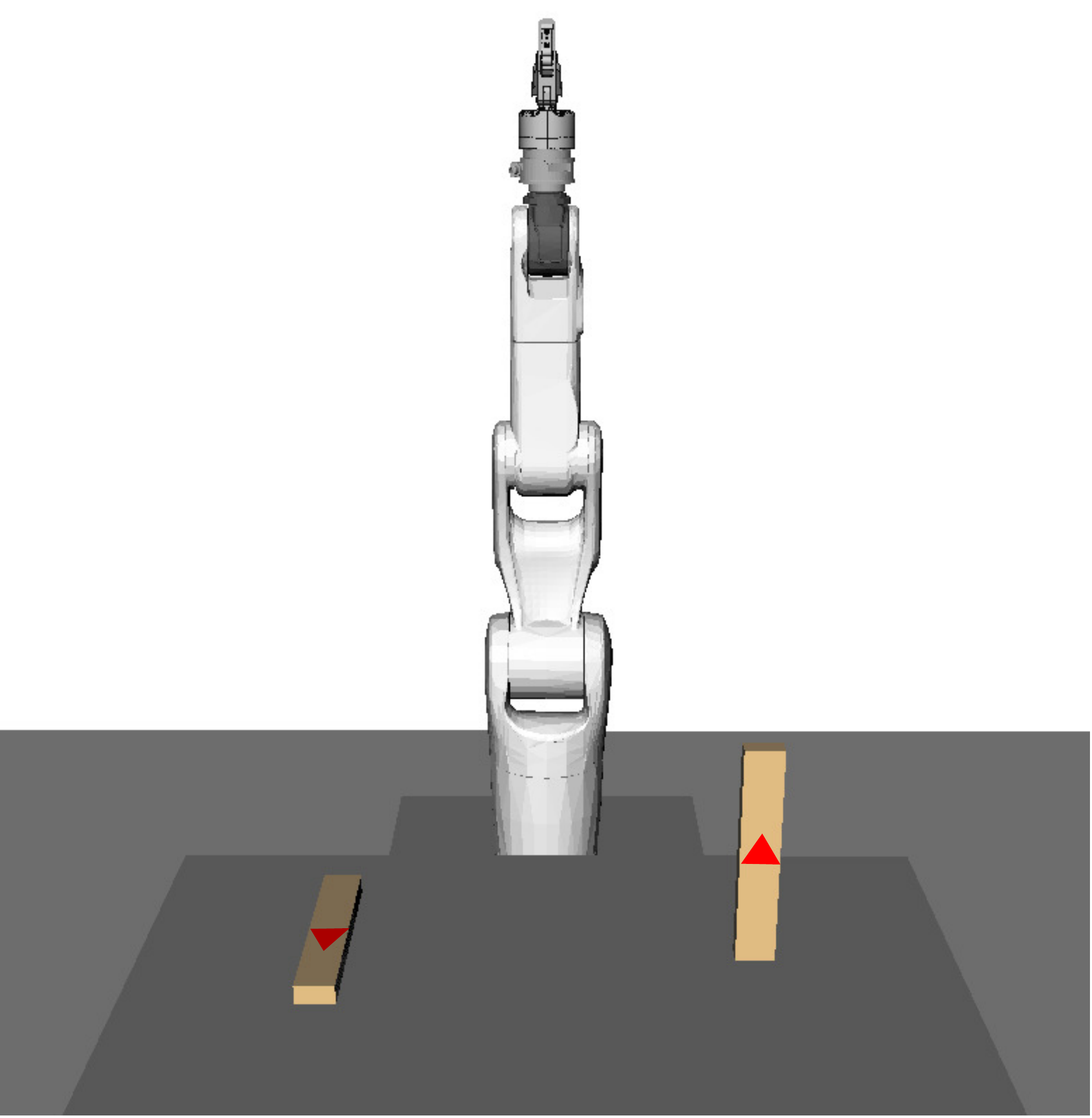}}
  \hspace{10pt}
  \subfloat[]{\includegraphics[width=0.13\textwidth]{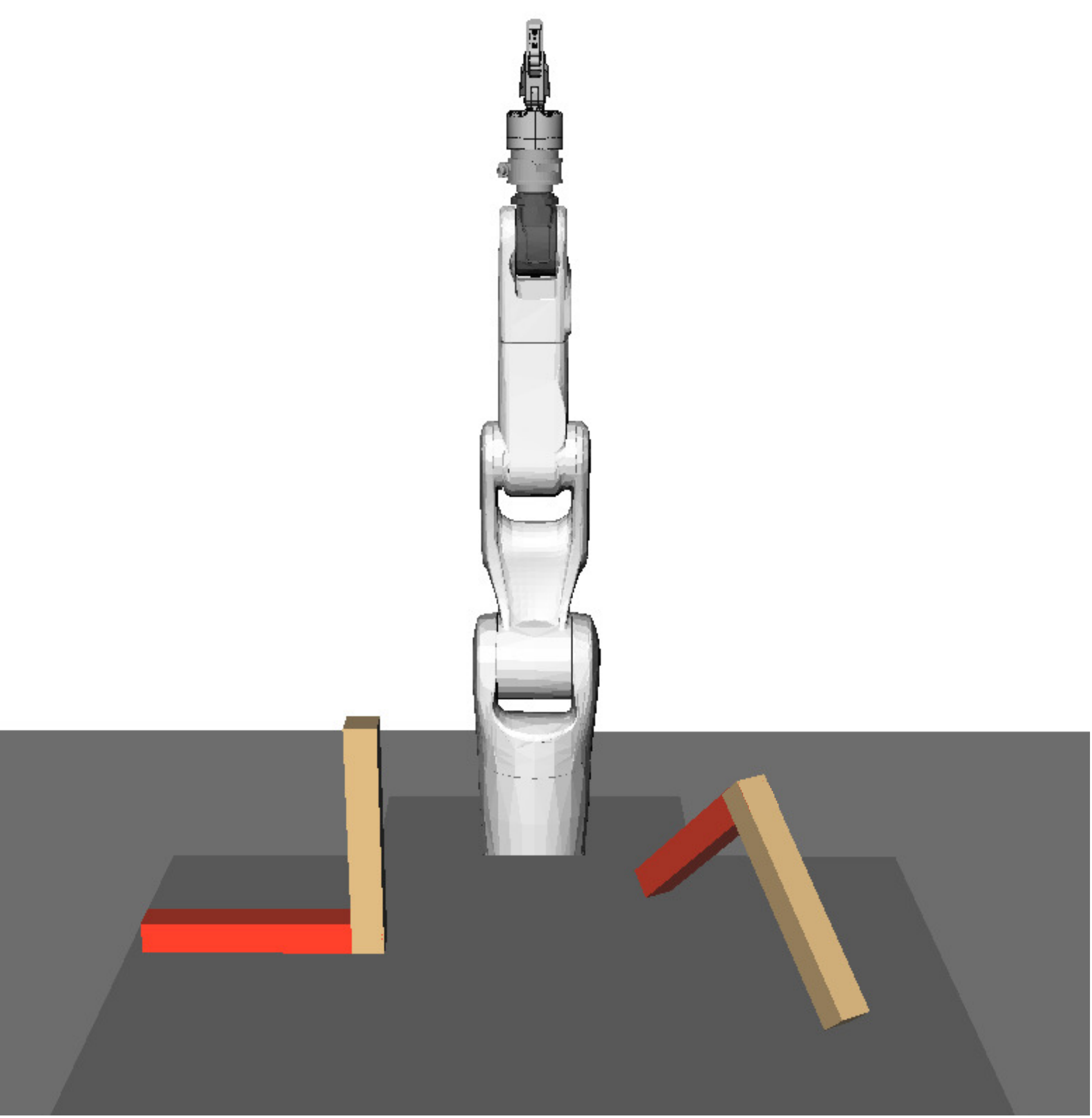}}
  \hspace{10pt}
  \subfloat[]{\includegraphics[width=0.13\textwidth]{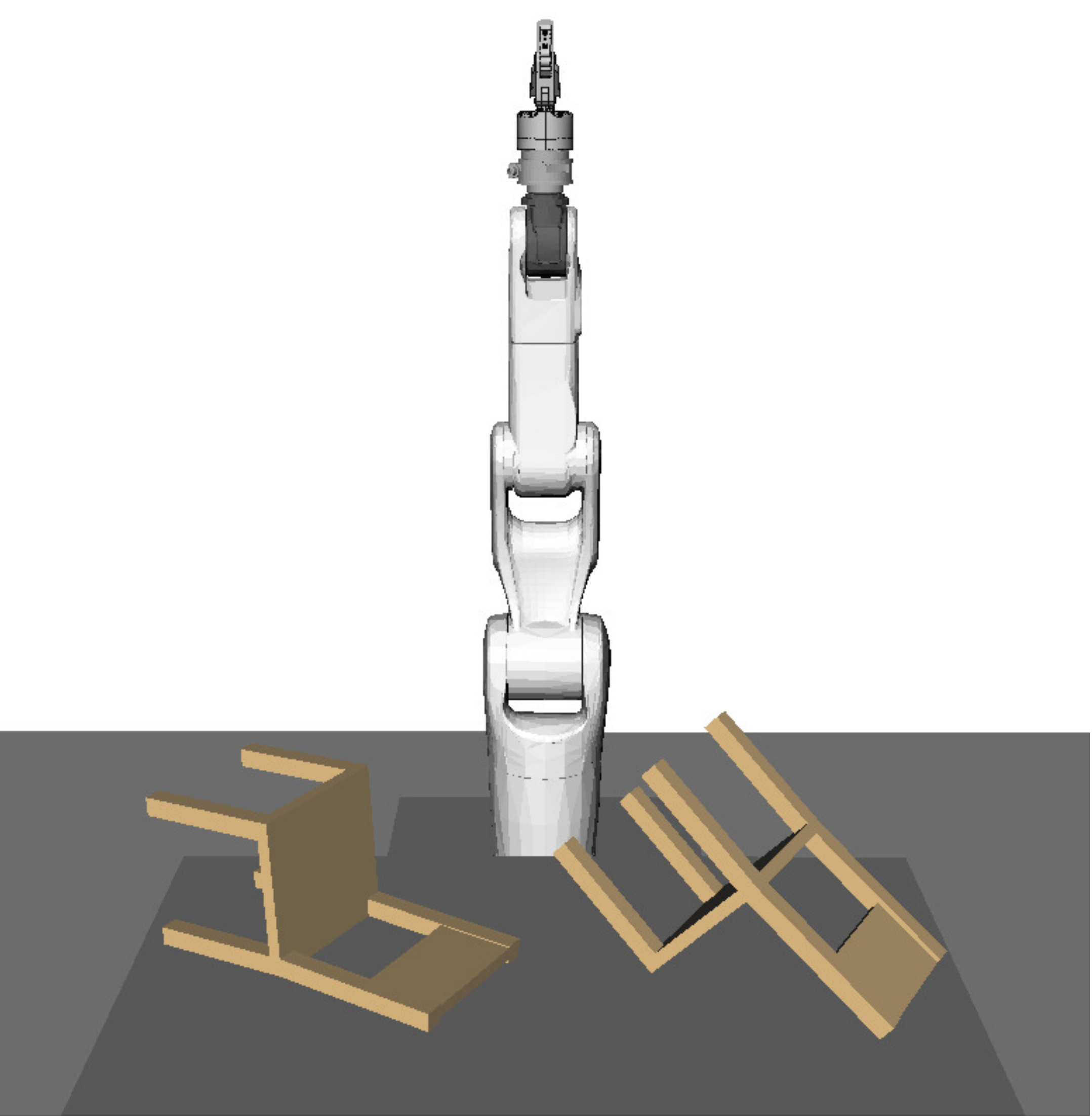}}
  \caption{Scenes used in our experiments. There are two identical
    (movable) objects shown in each figure. The objects on the left
    are at the initial transformations. The objects on the right are at
    the final transformations.}
  \label{figure:scenes}
\end{figure}

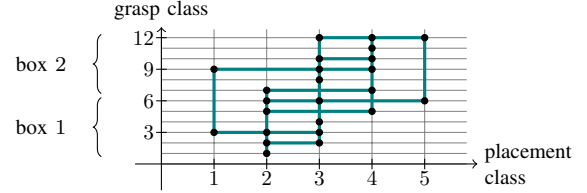
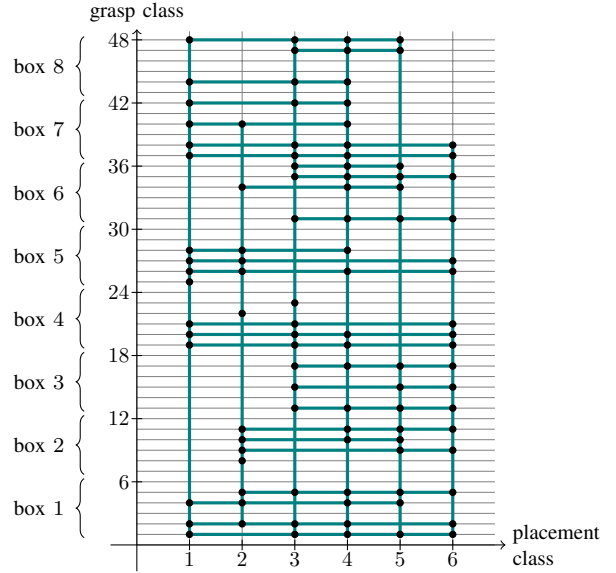
\begin{figure}
  \centering
  \subfloat[The high-level Grasp-Placement Table for an L-shaped object]{
    \begin{tikzpicture}[scale=0.7]
      \def\np{5}
      \def\ng{12}
      \def\ystep{0.2}
      \draw[ystep = \ystep cm, gray, ultra thin](0, 0) grid 
      (\np + 0.8, \ng*\ystep + 0.8*\ystep);
      \draw[->] (-0.5, 0) -- (\np + 1, 0);
      \draw[->] (0, -0.5) -- (0, \ng*\ystep + \ystep);
      \node [align=left, right, scale=0.8] at (\np + 1, 0) {placement\\
        class};
      \foreach \x in {1, ..., \np}{
        \draw (\x, -0.1) -- (\x, 0.1);
        \node [below, scale=0.8] at (\x, 0) {$\x$};
      }
      \node [align=left, above, scale=0.8] at (0, \ng*\ystep + \ystep)
      {grasp class};
      \foreach \y in {3, 6, ..., \ng}{
        \draw (-0.1, \y*\ystep) -- (0.1, \y*\ystep);
        \node [left, scale=0.8] at (0, \y*\ystep) {$\y$};
      }

      \draw[very thick, teal] (1, 3*\ystep) -- (1, 9*\ystep);
      \draw[very thick, teal] (2, 1*\ystep) -- (2, 7*\ystep);
      \draw[very thick, teal] (3, 2*\ystep) -- (3, 12*\ystep);
      \draw[very thick, teal] (4, 5*\ystep) -- (4, 12*\ystep);
      \draw[very thick, teal] (5, 6*\ystep) -- (5, 12*\ystep);
      \draw[very thick, teal] (2, 2*\ystep) -- (3, 2*\ystep);
      \draw[very thick, teal] (1, 3*\ystep) -- (3, 3*\ystep);
      \draw[very thick, teal] (2, 5*\ystep) -- (4, 5*\ystep);
      \draw[very thick, teal] (2, 6*\ystep) -- (5, 6*\ystep);
      \draw[very thick, teal] (2, 7*\ystep) -- (4, 7*\ystep);
      \draw[very thick, teal] (1, 9*\ystep) -- (4, 9*\ystep);
      \draw[very thick, teal] (3, 10*\ystep) -- (4, 10*\ystep);
      \draw[very thick, teal] (3, 12*\ystep) -- (5, 12*\ystep);
      \foreach \x/\y in 
      { 1/3, 1/9,
        2/1, 2/2, 2/3, 2/5, 2/6, 2/7,
        3/2, 3/3, 3/4, 3/6, 3/8, 3/9, 3/10, 3/12,
        4/5, 4/7, 4/9, 4/10, 4/11, 4/12,
        5/6, 5/12
      }{
        \filldraw (\x, \y*\ystep) circle [radius=0.06cm, fill=black];
      }
      \foreach \k in {1, 2}{
        \draw [decorate, decoration={brace, amplitude=3pt}, xshift=-4pt,
        yshift=0pt] (-1, 0.7*\ystep + 6*\k*\ystep - 6*\ystep) -- 
        (-1, 0.3*\ystep + 6*\k*\ystep) 
        node [black, midway, xshift=-0.8cm] {\footnotesize box $\k$};
      }
    \end{tikzpicture}
    \label{figure:graph_L}
  }
  
  \subfloat[The high-level Grasp-Placement Table for a small chair]{
    \begin{tikzpicture}[scale=0.7]
      \def\np{6}
      \def\ng{48}
      \def\ystep{0.2}
      \draw[ystep = \ystep cm, gray, ultra thin](0, 0) grid 
      (\np + 0.8, \ng*\ystep + 0.8*\ystep);
      \draw[->] (-0.5, 0) -- (\np + 1, 0);
      \draw[->] (0, -0.5) -- (0, \ng*\ystep + \ystep);
      \node [align=left, right, scale=0.8] at (\np + 1, 0) {placement\\
        class};
      \foreach \x in {1, ..., \np}{
        \draw (\x, -0.1) -- (\x, 0.1);
        \node [below, scale=0.8] at (\x, 0) {$\x$};
      }
      \node [align=left, above, scale=0.8] at (0, \ng*\ystep + \ystep)
      {grasp class};
      \foreach \y in {6, 12, ..., \ng}{
        \draw (-0.1, \y*\ystep) -- (0.1, \y*\ystep);
        \node [left, scale=0.8] at (0, \y*\ystep) {$\y$};
      }

      \foreach \x/\y/\z in {
        1/1/48, 2/2/40, 3/1/48, 4/1/48, 5/1/48, 6/1/38
      }{
        \draw[very thick, teal] (\x, \y*\ystep) -- (\x, \z*\ystep);
      }
      \foreach \x/\y/\z in {
        1/1/6, 2/1/6, 4/1/5, 5/2/6,
        9/2/6, 10/2/5, 11/2/6, 
        13/3/6, 15/3/6, 17/3/6,
        19/1/6, 20/1/6, 21/1/6, 
        26/1/6, 27/1/6, 28/1/4,
        31/3/6, 34/2/5, 35/3/6, 36/3/5,
        37/1/6, 38/1/6, 40/1/4, 42/1/4,
        44/1/4, 47/3/5, 48/1/5
      }{
        \draw[very thick, teal] (\y, \x*\ystep) -- (\z, \x*\ystep);
      }
      \def\radius{0.06}
      \foreach \y in {
        1, 2, 4, 19, 20, 21, 25, 26, 27, 28, 37, 38, 40, 42, 44, 48
      }
      {\filldraw (1, \y*\ystep) circle [radius=\radius cm, fill=black];}

      \foreach \y in {
        2, 4, 5, 8, 9, 10, 11, 22, 26, 27, 28, 34, 40
      }
      {\filldraw (2, \y*\ystep) circle [radius=\radius cm, fill=black];}

      \foreach \y in {
        1, 2, 5, 13, 15, 17, 19, 20, 21, 23, 31, 35, 36, 37, 38, 42, 44, 47, 48
      }
      {\filldraw (3, \y*\ystep) circle [radius=\radius cm, fill=black];}

      \foreach \y in {
        1, 2, 4, 5, 10, 11, 13, 17, 19, 20, 26, 28, 31, 34, 35, 36, 37, 
        38, 40, 42, 44, 47, 48
      }
      {\filldraw (4, \y*\ystep) circle [radius=\radius cm, fill=black];}

      \foreach \y in {
        1, 4, 5, 9, 10, 11, 13, 15, 17, 31, 34, 35, 36, 47, 48
      }
      {\filldraw (5, \y*\ystep) circle [radius=\radius cm, fill=black];}

      \foreach \y in {
        1, 2, 5, 9, 11, 13, 15, 17, 19, 20, 21, 26, 27, 31, 35, 37, 38
      }
      {\filldraw (6, \y*\ystep) circle [radius=\radius cm, fill=black];}

      brackets
      \foreach \k in {1, 2, ..., 8}{
        \draw [decorate, decoration={brace, amplitude=3pt}, xshift=0pt,
        yshift=0pt] (-1, 0.7*\ystep + 6*\k*\ystep - 6*\ystep) -- 
        (-1, 0.3*\ystep + 6*\k*\ystep) 
        node [black, midway, xshift=-0.6cm] {\footnotesize box $\k$};
      }
    \end{tikzpicture}
    \label{figure:graph_chair}
  }
  \caption{High-level Grasp-Placement Tables of objects used in our
    simulations.}
\end{figure}

The planners had to plan pick-and-place motions for three objects\,: a
box, an L-shaped object, and a small chair. The robot was to move from
\kw{Home}, \ie, $\qvect = \bm{0}$, pick up the object at a given
object transformation, place it at another given transformation, and
finally return to \kw{Home}. Snapshots of all three settings are shown
in Fig.~\ref{figure:scenes}. The Grasp-Placement Tables of the
L-shaped object and the chair are shown in Fig.~\ref{figure:graph_L}
and Fig.~\ref{figure:graph_chair}, respectively. (The Grasp-Placement
Table of the box is similar to the one in
Fig.~\ref{figure:graph_box}.)  These tasks are useful especially for
assembly operations, such as furniture assembly where the robot needs
to grasp a furniture part at some placement and move it to some
desired transformation to attach the part to other parts. We set chose
start and goal object transformations such that the robot needed to
perform regrasp operations at least once in order to complete the
tasks.

\captionsetup[subfigure]{position=b}

\begin{table*}
  \centering
  \caption{Pre-Processing time, planning time, and numbers of transitions from
    three problems averaged over $100$ runs.}
  \label{table:results}
  \renewcommand{\arraystretch}{1.5}
  \begin{tabularx}{1.00\textwidth}
    {
      @{\extracolsep{\fill}}|
      >{\setlength\hsize{0.2\hsize}\centering}X||
      >{\setlength\hsize{0.32\hsize}\centering}X|
      >{\setlength\hsize{0.32\hsize}\centering}X|
      >{\setlength\hsize{0.28\hsize}\centering}X|
      >{\setlength\hsize{0.23\hsize}\centering}X||
      >{\setlength\hsize{0.32\hsize}\centering}X|
      >{\setlength\hsize{0.32\hsize}\centering}X|
      >{\setlength\hsize{0.28\hsize}\centering}X|
      >{\setlength\hsize{0.23\hsize}\centering}X||
      >{\setlength\hsize{0.32\hsize}\centering}X|
      >{\setlength\hsize{0.32\hsize}\centering}X|
      >{\setlength\hsize{0.28\hsize}\centering}X|
      >{\setlength\hsize{0.23\hsize}\centering}X|
    }
    \cline{2-13}
    \multicolumn{1}{c|}{}
    & \multicolumn{4}{c||}{Problem $1$ (Box)}
    & \multicolumn{4}{c||}{Problem $2$ (L-shaped object)}
    & \multicolumn{4}{c|}{Problem $3$ (Small chair)}
    \tabularnewline
    \cline{2-13}
    \multicolumn{1}{c|}{}
    & prep. time (s.) & plan. time (s.) & \# transitions & 
    success rate
    & prep. time (s.) & plan. time (s.) & \# transitions & 
    success rate
    & prep. time (s.) & plan. time (s.) & \# transitions & 
    success rate
    \tabularnewline
    \hline
    \multicolumn{1}{|m{0.09\linewidth}|}{Our planner}
    & {$0.24$} & {$12.18$} & {$4.0$} & {$100 \%$}
    & {$0.50$} & {$15.32$} & {$4.0$} & {$100 \%$}
    & {$1.95$} & {$29.19$} & {$4.0$} & {$100 \%$}
    \tabularnewline
    \hline
    \multicolumn{1}{|m{0.09\linewidth}|}{PMP}
    & {--} & {$32.02$} & {$5.52$} & {$100 \%$}
    & {--} & {$46.83$} & {$5.07$} & {$73 \%$}
    & {--} & {$68.82$} & {$5.88$} & {$50 \%$}
    \tabularnewline
    \hline
    \multicolumn{1}{|m{0.09\linewidth}|}{DBMP}
    & {$35.53$} & {$17.99$} & {$4.0$} & {$72 \%$}
    & {$55.18$} & {$39.76$} & {$4.0$} & {$53 \%$}
    & {$102.38$} & {$37.08$} & {$4.0$} & {$46 \%$}
    \tabularnewline
    \hline
  \end{tabularx}
  \vspace{-10pt}
\end{table*}

\subsection{Descriptions of Alternative Manipulation Planners}

We implemented two alternative manipulation planners, each of which
lies on opposite ends of the spectrum of Manipulation Graph
construction. The first planner does not explicitly construct the
graph while the second constructs the graph by means of
discretization. Our planner lies midway between the two.

\subsubsection{Primitive Manipulation Planner (PMP)}
\label{subsubsection:pmp}

\begin{figure}
  \centering
  \begin{tikzpicture}[>=stealth', shorten >=1pt, node distance=2cm,
    auto, bend angle=40, every loop/.style={min
      distance=3mm, in=-20, out=20, looseness=5}, scale=0.8,
    inner sep=1pt, minimum size=1pt]
    \node[state] (p) {\small$\mc{P}$};
    \node[state] (gp) [right = of p]{\small$\GP$};

    \path[->]
    (p) edge [bend left] node [anchor=south, scale=0.8] {transit} (gp)
    (gp) edge [bend left] node [anchor=south, scale=0.8] {transit} (p)
    (gp) edge [loop right] node [anchor=west, scale=0.8, align=left] 
    {transit $\rightarrow$ transfer or \\ transfer $\rightarrow$ transit} (gp);
  \end{tikzpicture}
  \caption{A transition diagram used in Primitive Manipulation Planner.}
  \label{figure:transition}
\end{figure}
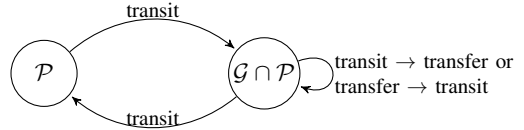

PMP has minimal knowledge of the structure of $\mc{C}$ as it has no
pre-processing stage. It explores $\mc{C}$ according to the transition
diagram, similar to~\cite{LM15hal}, shown in
Fig.~\ref{figure:transition}.

We implemented PMP as a bidirectional tree-based planner. It grows one
tree rooted at $(\q{start}, \T{start})$, the other rooted at
$(\q{goal}, \T{goal})$. In each iteration, it samples a new composite
configuration and tries to connect it with existing vertices on a
tree. For example, if the newly sampled configuration is in $\GP$ and
is to be connected with a configuration in $\mc{P}$ on a tree, then
the local planner will try connecting them with a transit path. For
the distance metric used in our implementation, we defined a distance
$d$ between two configurations $(\qvect_1, \Tvect_1)$ and $(\qvect_2,
\Tvect_2)$ as a weighted sum of the Euclidean distance between the
robot configurations and a distance between the object
transformations, \ie, $d = \alpha\Vert \qvect_2 - \qvect_1 \Vert^2 +
(1 - \alpha) w(\Tvect_1, \Tvect_2)$, where $0 \leq \alpha \leq 1$ and
$w(\Tvect_1, \Tvect_2)$ is a weighted sum of the minimal geodesic
distance between two rotations (see~\cite{PR97tg} for more detail) and
the Euclidean distance between two displacements.

\subsubsection{Discretization-Based Manipulation Planner (DBMP)}
\label{subsubsection:dbmp}
In contrast with PMP, DBMP constructs its variant of Manipulation
Graph, \emph{two-layer regrasp graph}~\cite{WanX15icra}, by
discretizing placements and grasps

We implemented DBMP following~\cite{WanX15icra}. The planner starts by
constructing the set of possible grasps by means of sampling. Then it
builds a two-layer regrasp graph\,: the first layer composes of
placements and the second layer composes of grasps. Two placements in
the first layer are connected together if they share at least one
common valid grasp. To solve a query, DBMP first searches for
placement sequences that connect the start and goal placements. Then
it examines each placement sequence and searches for feasible grasps
associated with it.

However, since all the searches are deterministic and proceed in a
depth-first fashion, the time-complexity of DBMP is significantly
large. Instead of having three varying parameters for each placement
class\,: two parameters for a location on the table and one parameter
for the rotation about an axis normal to the table's surface, we
constrained each placement class to have only a single varying
parameter, the rotation. The same was also done implicitly
in~\cite{WanX15icra}. Furthermore, the deterministic search is also
much sensitive to indexing of grasps and placements. To reduce this
effect we shuffled grasp and placement indices before each run.

Note that our implementation of grasp set computation was different
from the method used in~\cite{WanX15icra}, which was basically a
discretization. Therefore, the pre-processing time put in
Table~\ref{table:results} is intended only for reference. The numbers
of total grasps computed in the cases of a box, an L-shaped object,
and a chair are $124$, $242$, and $331$, respectively.

\subsection{Simulation Results and Comparisons}

For each object, we ran each of the three planners $100$ times. The
data collected are pre-processing time, planning time, numbers of
transitions, and success rate. (If no solution was found in $100$ s.,
then the run was considered failed.) Data were averaged over
successful runs and reported in Table~\ref{table:results}.

\subsubsection{Comparison with Primitive Manipulation Planner}
From the data reported in Table~\ref{table:results}, we can see that
by exploiting more information about the connectivity of subsets of
$\GP$, \ie, by constructing high-level Grasp-Placement Tables, our
planner was able to search for manipulation paths more systematically
and efficiently, as reflected through the running time and the path
quality achieved by our planner. As PMP had no information about the
connectivity, much planning time was spent attempting infeasible
connections between $\GP$ configurations. Furthermore, when the object
was composed of more boxes, connection attempts were even more prone
to failure since from any subset of $\GP$, the ratio of the number of
reachable $\GP$ subsets to the number of all the subsets decreased.

\subsubsection{Comparison with Discretization-Based Manipulation
  Planner}
In fact, the two-layer regrasp graph~\cite{WanX15icra} is similar to
ours. One major difference is that they did not exploit any grasp
parameterization in building the graph. Therefore, in their case, each
grasp class contains exactly one grasp, hence numerous grasp
classes. Also, the deterministic search employed in DBMP makes its
capability relatively limited. Firstly, all placements have to be
constrained to only one location on the table, or at most a few, due
to time complexity of the search. Second, a significant amount of time
is spent on exploring infeasible placement and grasp sequences due to
the depth-first fashion of the search. This makes the success rate of
DBMP in the given time much lower than the other planners.

\subsection{Hardware Experiment}
\begin{figure*}
  \centering
  \subfloat[{}]{\includegraphics[height=0.13\textwidth]{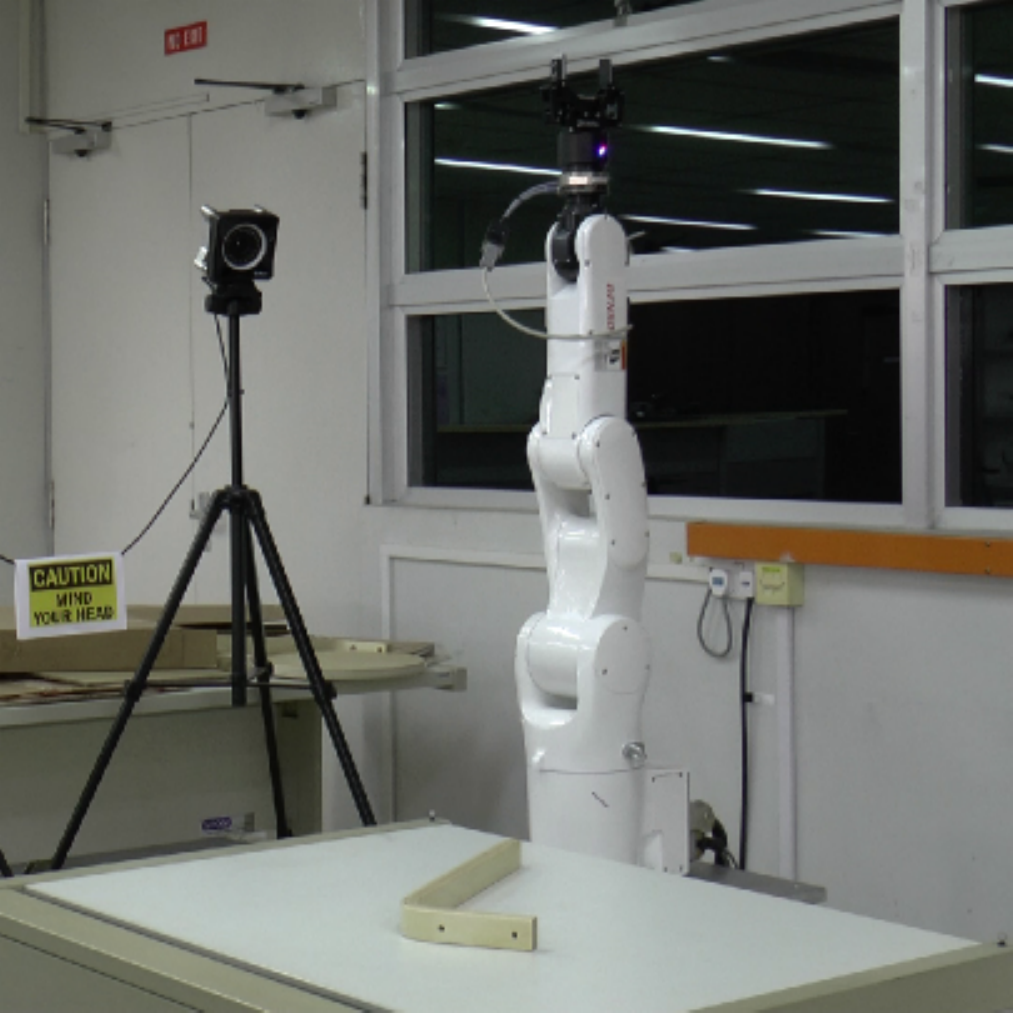}}
  \hspace{3pt}
  \subfloat[{}]{\includegraphics[height=0.13\textwidth]{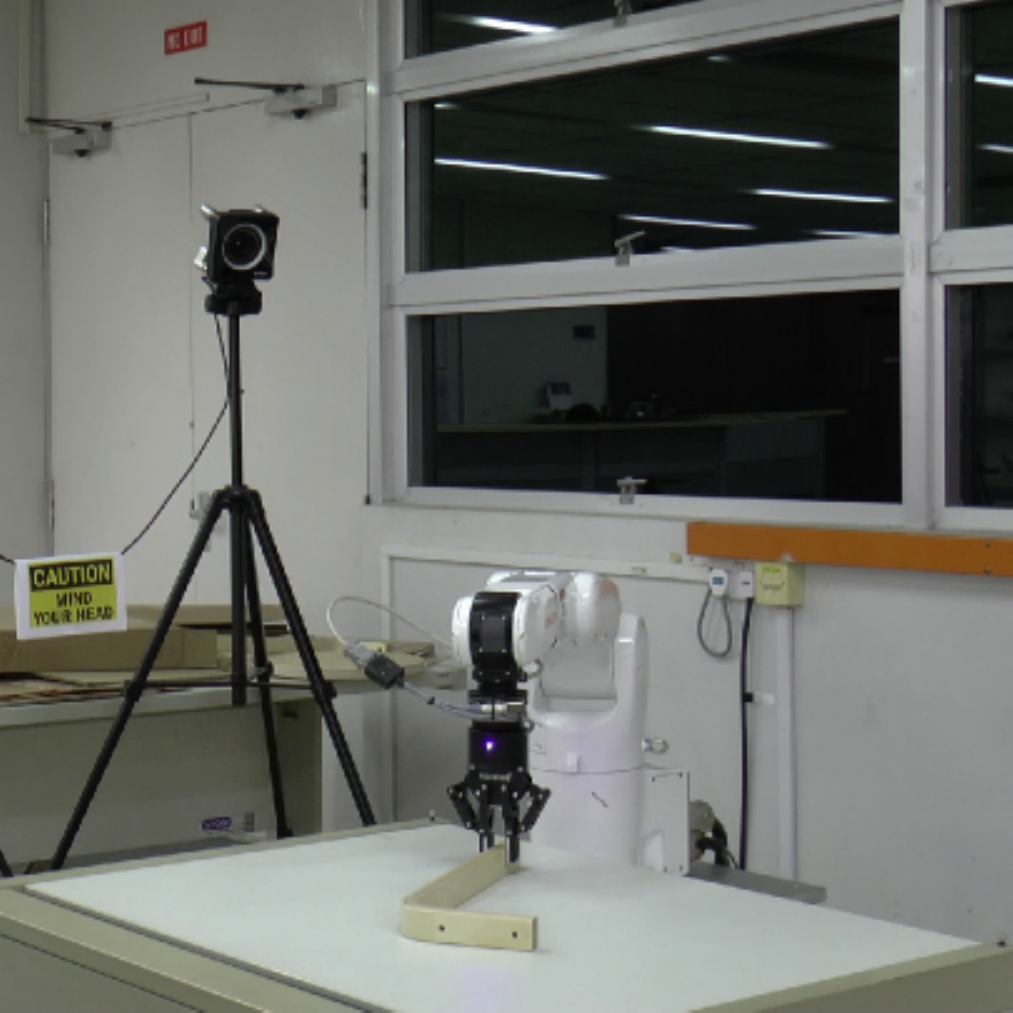}}
  \hspace{3pt}
  \subfloat[{}]{\includegraphics[height=0.13\textwidth]{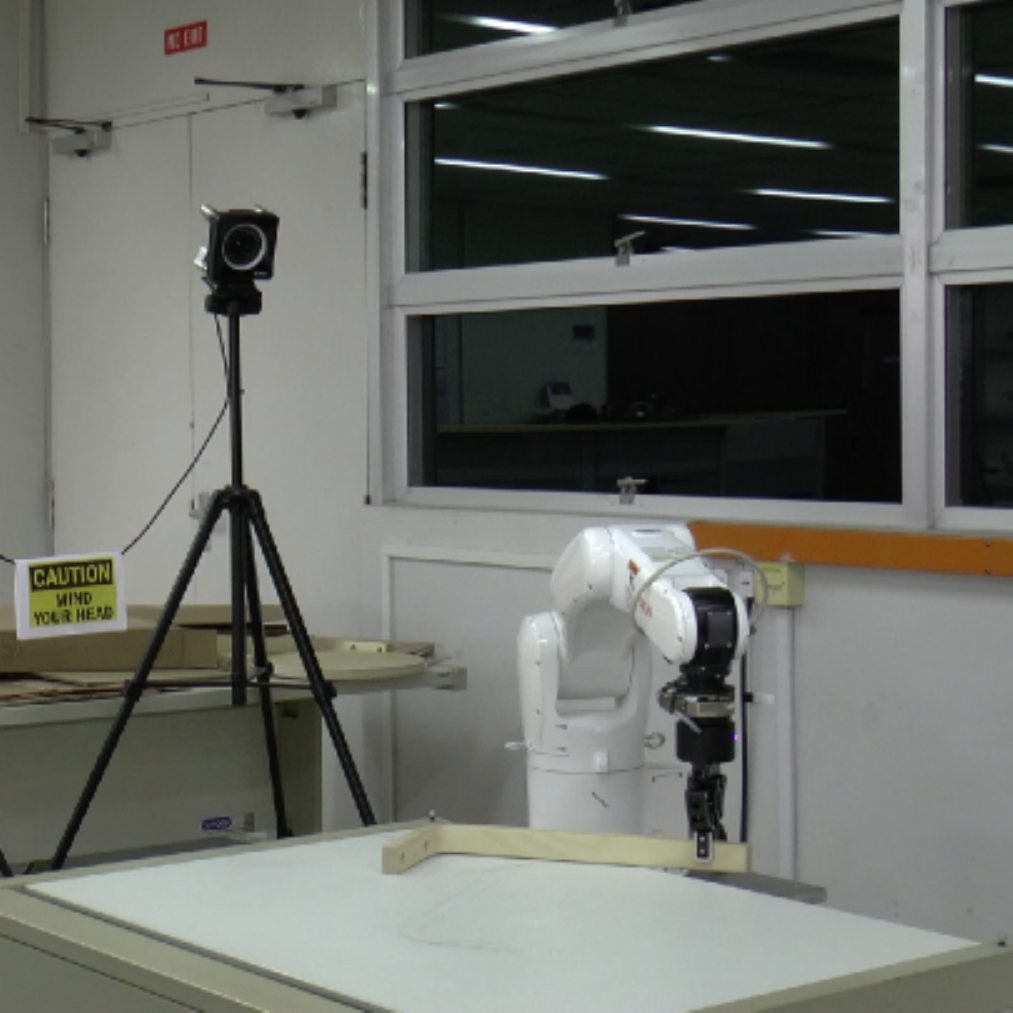}}
  \hspace{3pt}
  \subfloat[{}]{\includegraphics[height=0.13\textwidth]{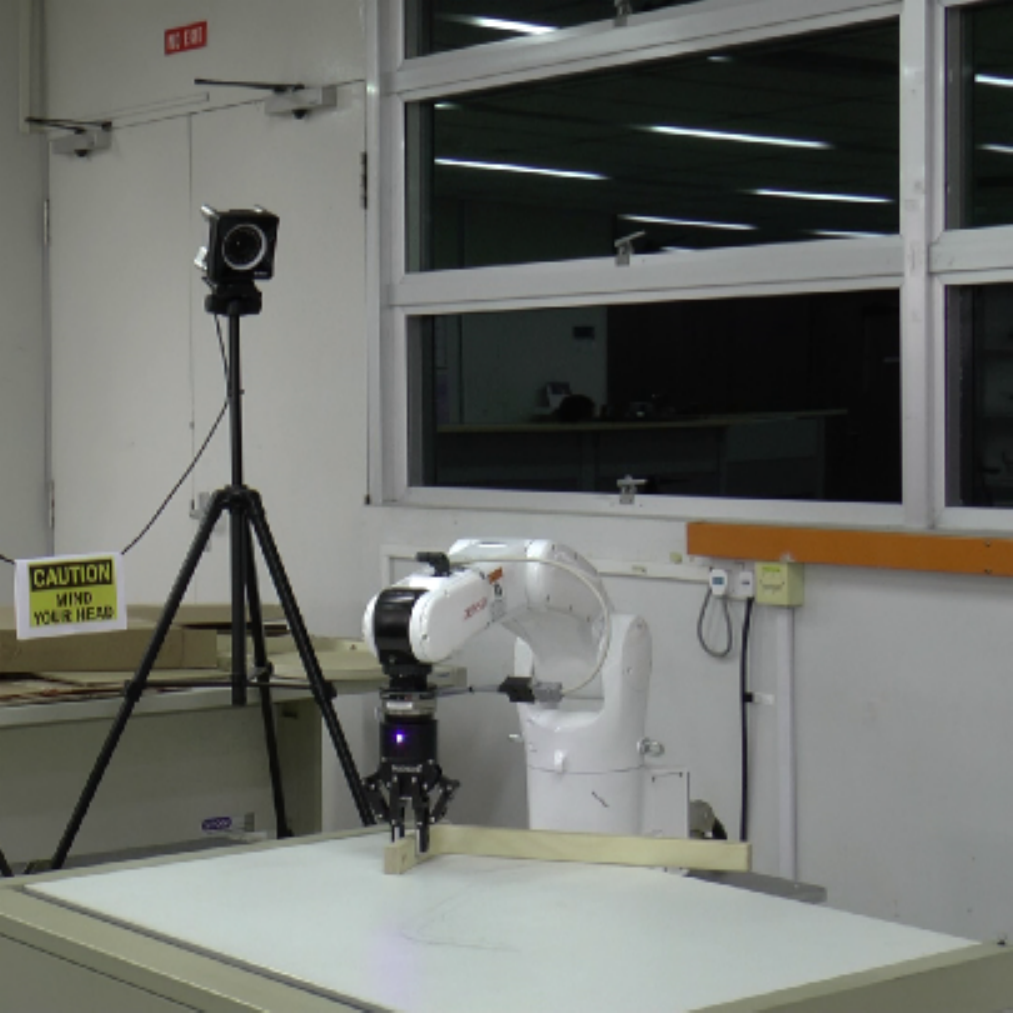}}
  \hspace{3pt}
  \subfloat[{}]{\includegraphics[height=0.13\textwidth]{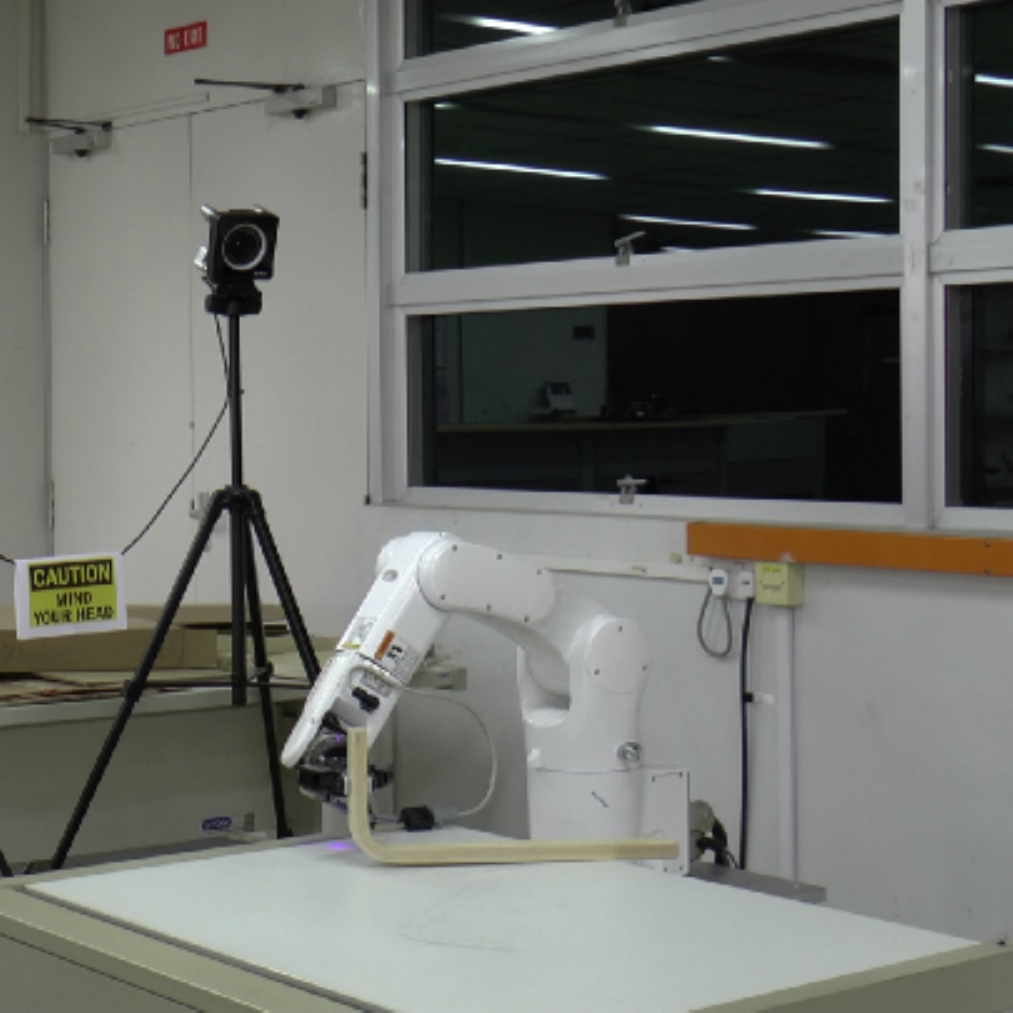}}
  \hspace{3pt}
  \subfloat[{}]{\includegraphics[height=0.13\textwidth]{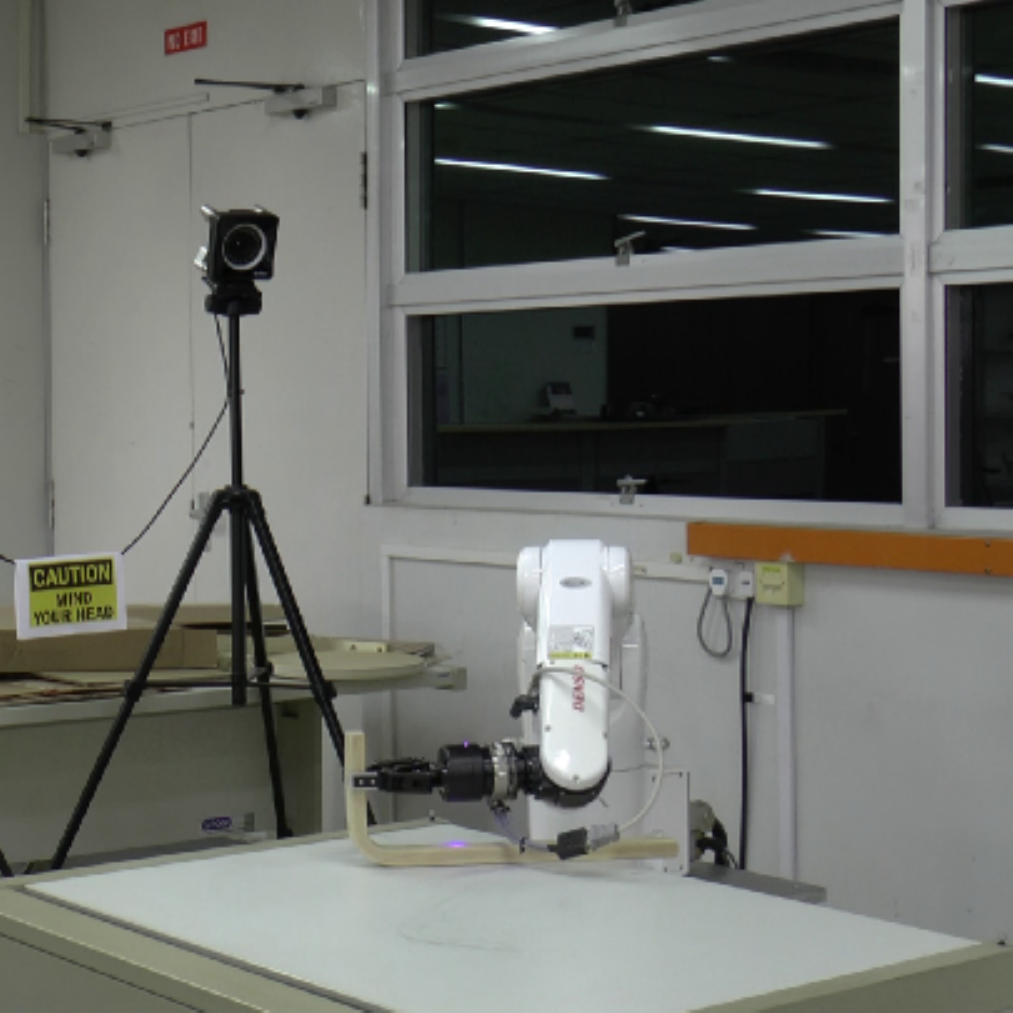}}
  \hspace{3pt}
  \subfloat[{}]{\includegraphics[height=0.13\textwidth]{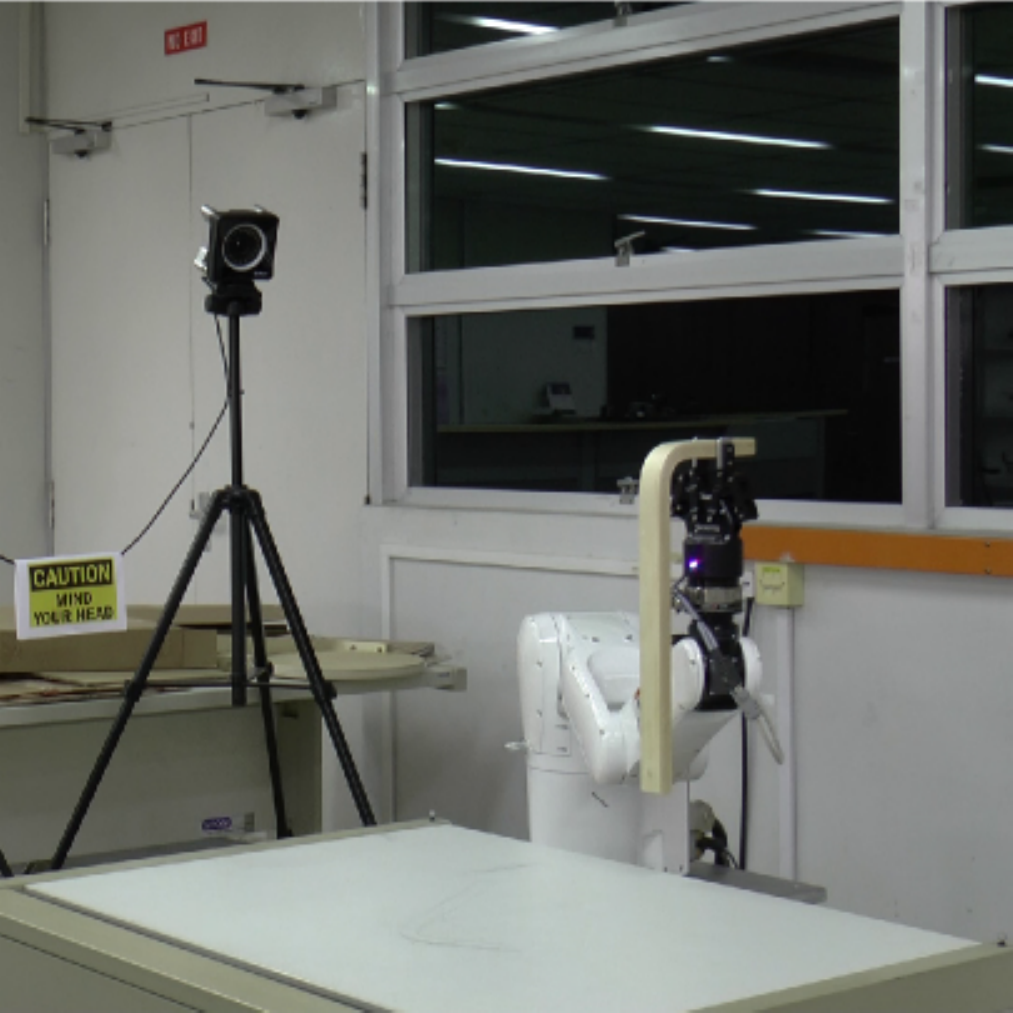}}
  \caption{Snapshots from the experiment on the real robot. (a) The
    start configuration. (b)--(f) The robot performed pick-and-place
    motions to move the object to a stable placement which allowed
    grasping with the desired final grasp. (g) The goal
    configuration. The video can be found at
    \tt{https://youtu.be/tLouwj0wITQ}.}
  \label{figure:realexp}
  \vspace{-15pt}
\end{figure*}

Besides simulations, we also conducted a hardware experiment. In this
experiment, the robot must pick up the leg of a stool, which we
approximated by three boxes, and hold it with a given grasp to
facilitate a subsequent screwing operation (to be performed by another
robot; not included in this experiment). Snapshots of the start and
goal configurations are shown in Fig.~\ref{figure:realexp}. One can
see that the initial pose of the object does not allow grasping with
the desired grasp. Thus, the robot needs to regrasp the object several
times. A video of the robot performing the task can be found at
{\tt{https://youtu.be/tLouwj0wITQ}}.

\captionsetup*[subfigure]{position = bottom}

\section{Discussion and Conclusion}
\label{section:discussion}

\subsection{Extension to Broader Classes of Object Models}
\label{subsection:extension}
Our method for constructing the high-level Grasp-Placement Table is
mainly based on parameterizations of grasp and placement
classes. Parameterization enables categorizing infinitely many grasps
and placements into a finite number of grasp and placement
\emph{classes}. This enables us to effectively capture the
connectivity of $\GP$ subsets and encode it into a high-level
Grasp-Placement Table without using any discretization.

Our method can generalize to models for which it is possible to find
efficient placement and grasp parameterizations. In general, it is not
difficult to find \emph{placement classes} of a general object as one
can compute the convex hull of the object and then test which surfaces
of the hull result in stable placements. Therefore, the main
requirement is that the object is \emph{grasp-parameterizable}.

\begin{figure}
  \centering
  \subfloat[]{\includegraphics[width=0.1\textwidth]{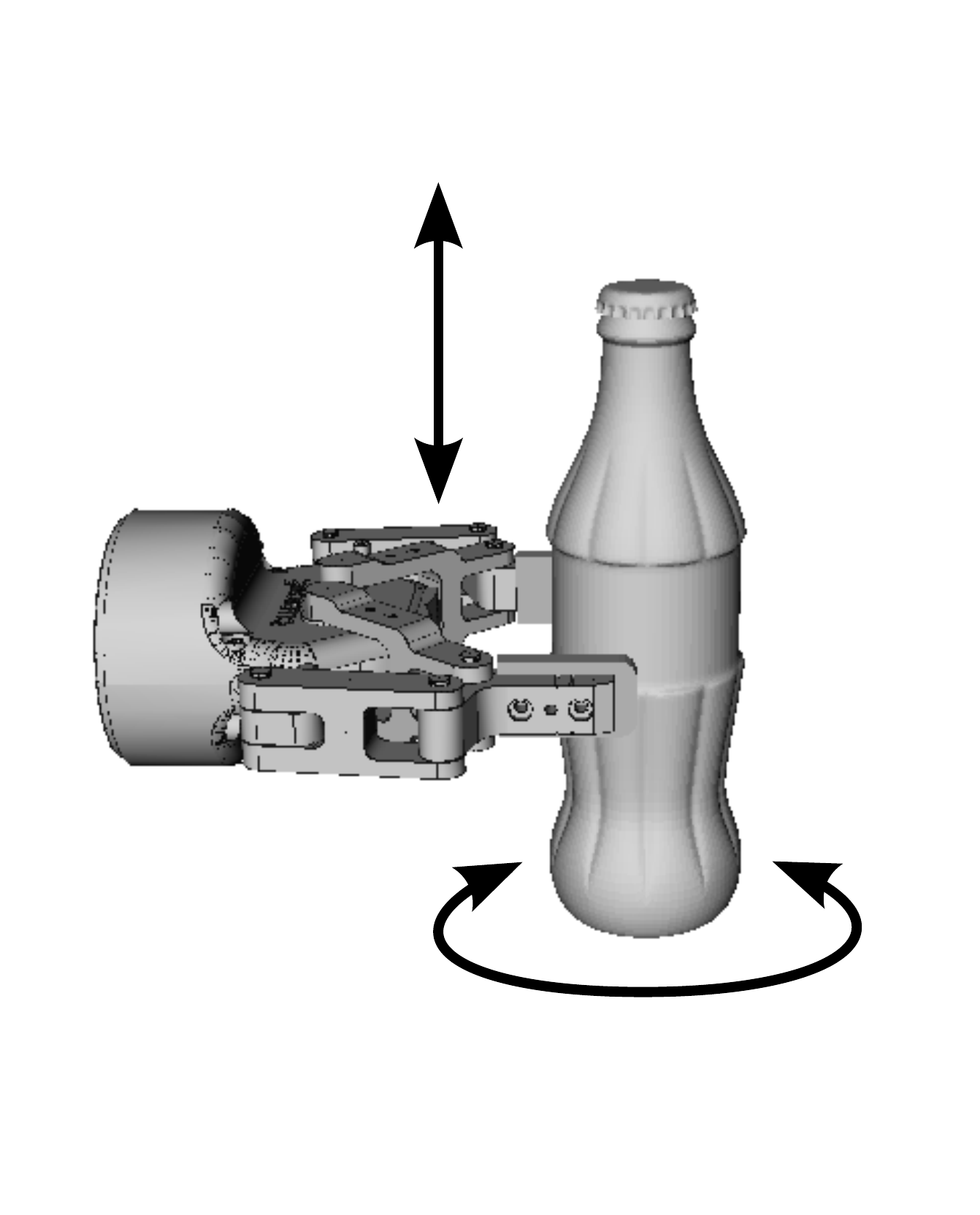}}
  \hspace{18pt}
  \subfloat[]{\includegraphics[width=0.1\textwidth]{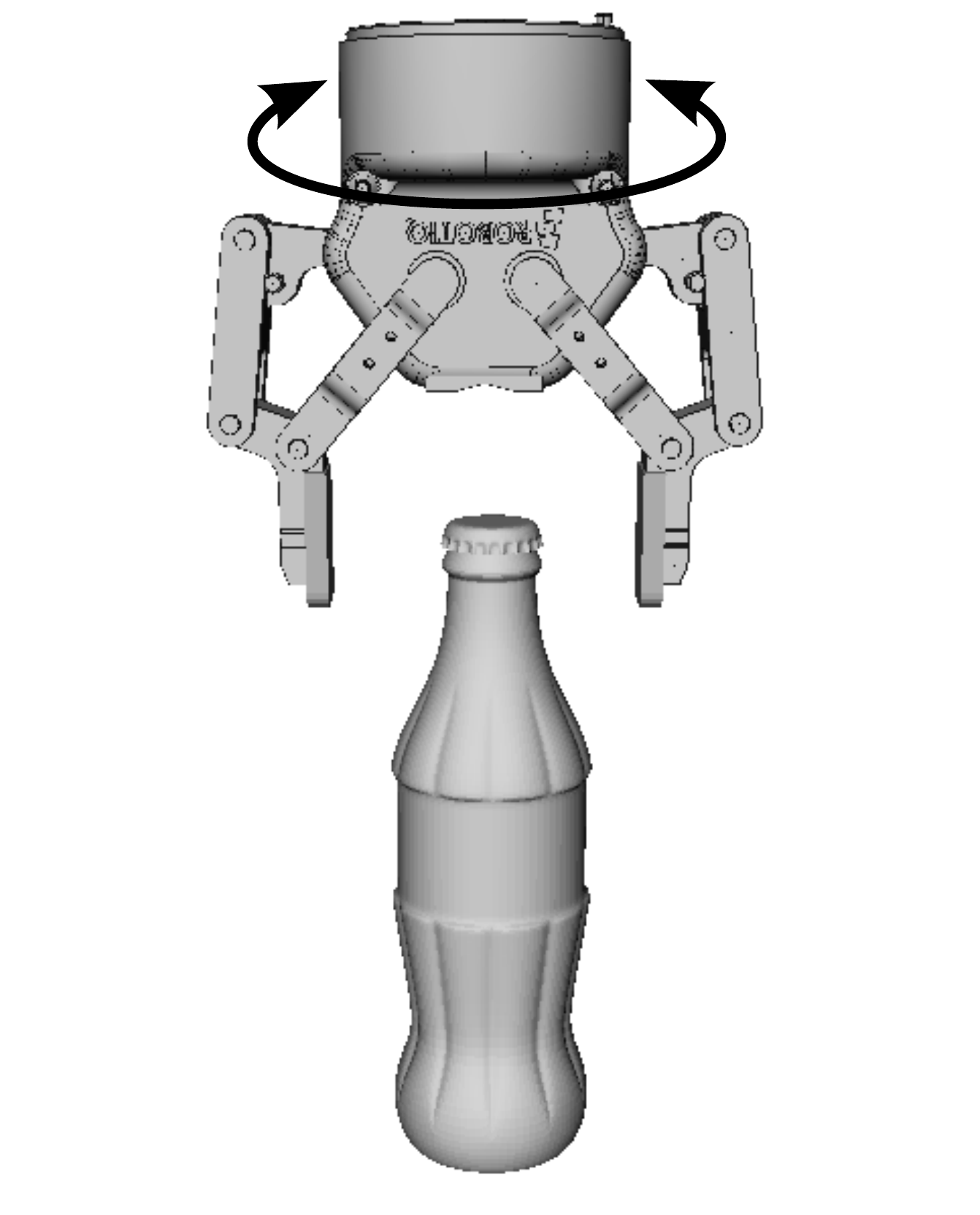}}
  \hspace{18pt}
  \subfloat[]{\includegraphics[width=0.1\textwidth]{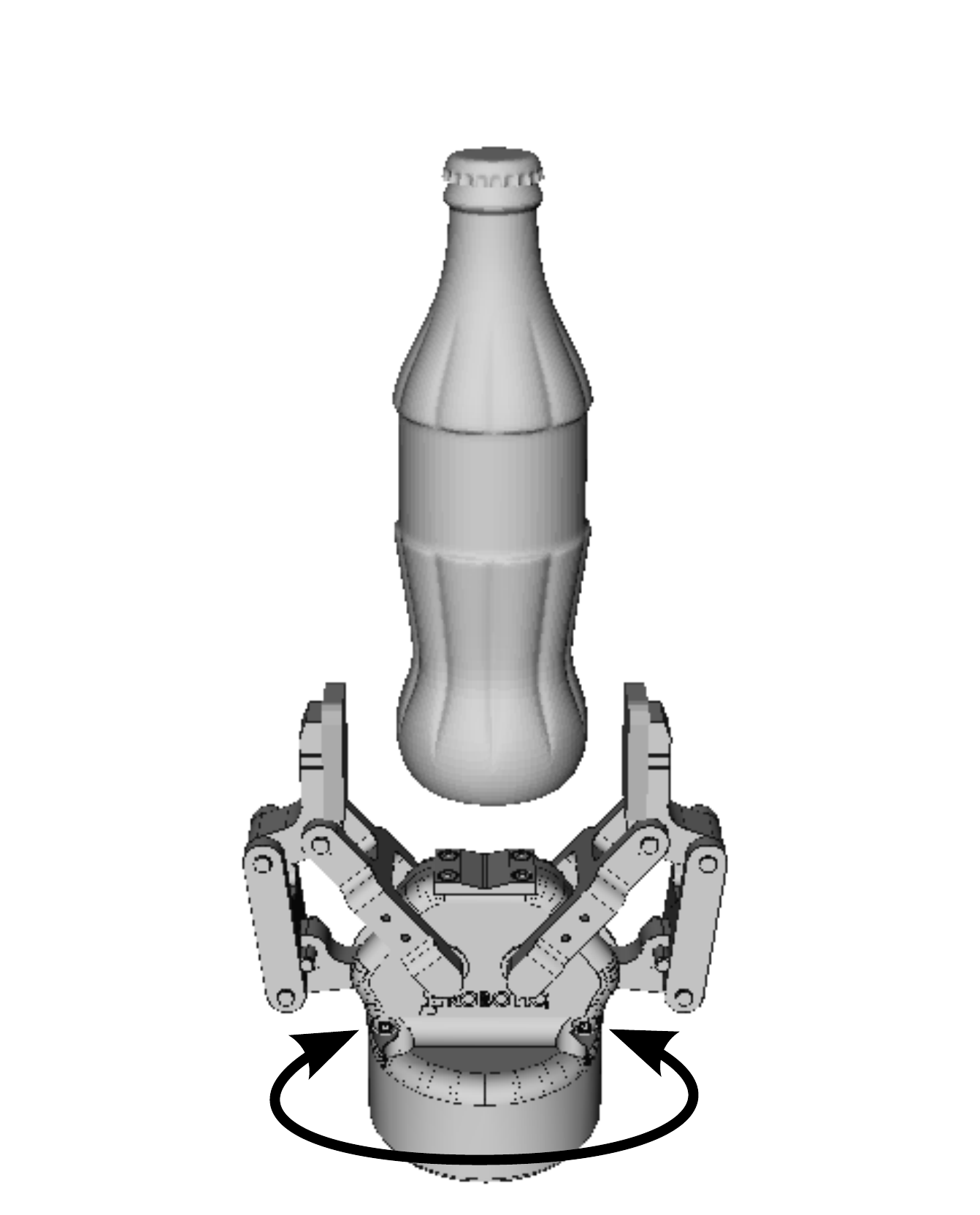}}
  \caption{Three possible grasp classes for a parallel gripper
    grasping a bottle. Arrows show gripper motions that result in
    grasps contained in the same class.}
  \label{figure:bottle}
\end{figure}

A wide variety of objects can indeed be grasp-parameterized, including
many daily-life objects. For example, for a bottle, one can categorize
grasps into three classes, see Fig.~\ref{figure:bottle}. For highly
irregular objects with no efficient grasp parameterization, one may
have to resort to grasp discretization. In such a case, our high-level
Grasp-Placement Table will be similar to the regrasp graph
of~\cite{WanX15icra}.

\subsection{Conclusion}
We have presented a manipulation planner to tackle pick-and-place
planning problems. We first proposed a method to construct a
high-level Grasp-Placement Table based on the models of the robot, the
object, and the environment, without resorting to discretization of
$\GP$, the fundamental set of configurations where transition between
transit and transfer paths may occur. Our construction is therefore
associated with a full parameterization of $\GP$, in contrast to
previously proposed methods. The Table then serves as a guide for the
planner to explore the composite configuration space.

Our method to construct the Grasp-Placement Table readily applies to
movable objects that are boxes or composed of boxes. In such cases,
assuming that the gripper can exert large enough forces with its
fingers, all grasp classes considered here are also force-closure. We
also discussed extension of the method to handle broader classes of
objects.

The experimental results presented in Section~\ref{section:results}
confirmed that the high-level Grasp-Placement Table helps improve both
running time and manipulation path quality as compared to existing
manipulation planners.

\bibliographystyle{IEEEtran} 
\bibliography{../../CRI_jr}

\end{document}